\RequirePackage[svgnames,table]{xcolor}

\documentclass[11pt,letterpaper,logo,twocolumn]{yalearxiv}
\usepackage{graphicx}
\usepackage{float,epstopdf}
\usepackage{bbm}

\usepackage{microtype}

\usepackage{natbib}
\setcitestyle{square}

\usepackage{subcaption}
\usepackage{booktabs}

\usepackage{amsmath}
\usepackage{amssymb}
\usepackage{mathtools}
\usepackage{amsthm}
\usepackage{dsfont}
\usepackage{multicol}
\usepackage{makecell}
\usepackage{multirow} 
\usepackage{amsfonts} 
\usepackage{mathrsfs}
\usepackage[amssymb, thickqspace]{SIunits}
\usepackage{enumitem}
\usepackage{pgfplotstable}
\usepackage{lipsum}		% Can be removed after putting your text content

\usepackage{placeins}

\usepackage{microtype}
\usepackage{graphicx}
\usepackage{booktabs} % for professional tables
\usepackage[table]{xcolor}
\usepackage{arydshln}
\usepackage[normalem]{ulem} % [normalem] prevents the package from changing the default behavior of `\emph` to underline.

\usepackage{cases}
\usepackage{wrapfig}

\usepackage{url}

\usepackage{thmtools}
\usepackage{thm-restate}
\usepackage{tabu}

\definecolor{huskypurple}{HTML}{4B2E83}

\usepackage{titletoc}

\usepackage{listings}
\lstdefinestyle{promptstyle}{
  basicstyle=\ttfamily\footnotesize,
  breaklines=true,
  breakautoindent=false,
  breakindent=0pt,
  postbreak=\mbox{\textcolor{gray}{$\hookrightarrow$}\space},
  columns=fullflexible,
  keepspaces=true,
  frame=single,
  framesep=5pt,
  xleftmargin=6pt,
  xrightmargin=6pt,
  aboveskip=8pt,
  belowskip=8pt,
  showstringspaces=false,
}

\newcommand{\pifive}{$\pi_{0.5}$}
\newcommand{\bench}{LoopQ}

\usepackage{cleveref}
\usepackage{pifont}

\usepackage[table]{xcolor}
\definecolor{grp}{RGB}{235,235,235}

\newcommand{\cmark}{\ding{51}}
\newcommand{\xmark}{\ding{55}}

\newcommand{\takeaway}[1]{%
  \par\noindent\textbf{Key takeaway:} #1\par
}

\newcommand{\grouprow}[1]{%
  \multicolumn{5}{l}{\textbf{#1}}\\
}

\makeatletter
\def\munderbar#1{\underline{\sbox\tw@{$#1$}\dp\tw@\z@\box\tw@}}
\makeatother

\AddToHook{cmd/appendix/before}{%
  \setcounter{axiom}{0}%
}

\newcommand{\be}{\begin{equation}}
\newcommand{\ee}{\end{equation}}
\newcommand{\bea}{\begin{equation*}\begin{aligned}}
\newcommand{\eea}{\end{aligned}\end{equation*}}

\newtcolorbox{simpleElegantQuote}{
    colback=AliceBlue!50!White,
    colframe=RoyalBlue!75!Black,
    boxrule=0.5pt,
    arc=2mm,
    boxsep=4pt,
    left=10pt, right=10pt,
    top=8pt, bottom=8pt,
    fontupper=\itshape,
}

\title{LoopQ: Fair, Closed-Loop Evaluation of Quantized Vision-Language-Action Models}

\runningtitle{LoopQ: Fair, Closed-Loop Evaluation of Quantized Vision-Language-Action Models}


\usepackage{fontawesome5}   % brand/utility icons, e.g. \faGithub

\definecolor{yaleblue}{RGB}{0,58,112}

\author{%
Syed Umer Bukhari, Magnus Almgren, Ahmed Ali-Eldin \\

Chalmers University of Technology and University of Gothenburg\\
Computer and Network Systems
}

\hypersetup{colorlinks=true, linkcolor=blue!50!black, citecolor=blue!50!black,
            urlcolor=blue!50!black}

\definecolor{grp}{gray}{0.92}

\renewcommand{\grouprow}[2]{%
  \rowcolor{grp}%
  \multicolumn{#1}{@{}l}{\textit{#2}}\\%
}
\providecommand{\sig}{\ensuremath{^{\ast}}}

\begin{document}

% -----------------------------------------------------------------------------
%  Abstract. In this template the abstract is *captured* here and then typeset
%  inside the title box by \maketitle, so it must appear BEFORE \maketitle.
%  The class adds the "Abstract" heading automatically. Keep the body to a
%  single paragraph.
% -----------------------------------------------------------------------------

%%%%%%%%%%%

\begin{abstract}

Post-training quantization (PTQ) for vision-language-action (VLA) models promises smaller and faster VLA deployments. However, when analyzing existing VLA quantization works, we realised that it is hard to reason about the performance of the suggested quantization method compared to existing, similar, or in many cases, simpler PTQ methods. We present \bench{}, a closed-loop benchmark that compares multiple quantization policies on the same checkpoint, layers and engine, varying only the method and the kernels. We evaluate RTN, AWQ, GPTQ, ActQuant and QuantVLA on four VLAs over more than 180{,}000 episodes on LIBERO and LIBERO-plus, compare against QVLA's reported results on the same layers, and audit whether four public VLA-quantization releases can be reproduced. We benchmark the methods on an NVIDIA H100, an L4, and a Jetson AGX Thor. Our results show that while quantization reliably saves memory, $2$--$3\times$ for real low-bit models, but VLA-specific PTQ methods underperform compared to simpler generic baselines. At 4 bits, RTN, AWQ and GPTQ match or exceed ActQuant and QuantVLA, plain RTN matches the calibrated methods, and plain AWQ matches the success QVLA reports for itself. Standard LIBERO hides the losses that do exist: configurations with no significant LIBERO loss lose 1.5--4.3 points under paired LIBERO-plus perturbations. Speed is a weaker story. At batch 1 on the H100, no quantized \pifive{} or OpenVLA-OFT configuration is faster than its 16-bit model ($1.06$--$1.58\times$ slower on library kernels); only the L4 with int8 kernels gains (4--14\%), the Thor is slower on every 4-bit configuration, and the engine alone changes latency by $2.1\times$. Profiling explains why: \pifive{}'s quantized backbone is compute-bound, while its memory-bound action expert, 63\% of GPU time and 13\% of the parameters, stays in 16 bits. Only one of the four releases can be timed. Smaller models and saturated benchmarks do not mean faster or more reliable robots.

\end{abstract}

\maketitle

\section{Introduction}
\label{sec:intro}
    Vision-language-action (VLA) models map camera images and a language instruction to robot actions by coupling a pretrained vision-language model (VLM) with an action decoder~\citep{kim2024openvla,black2024pi0,black2025pi05,nvidia2025gr00t,bu2025univla}. Their millions to billions of parameter backbones are hard to fit within the memory, latency and power budgets of robots, so post-training quantization (PTQ)~\citep{frantar2023gptq,lin2024awq,zhao2024atom,lin2025qserve}, is increasingly treated as a prerequisite for deployment. Recenlty, a large number of VLA-specific PTQ methods have claimed that action-aware sensitivity, mixed precision or activation smoothing yields smaller and faster policies without loss of task success~\citep{akbari2026actquant,zhang2026quantvla,xu2026qvla,wang2026holoq}.However, we find that these works rarely answer three basic questions about these claims.

\textbf{Is the gain over a fair baseline?}
VLA-specific methods are compared with generic methods such as AWQ~\citep{lin2024awq} and GPTQ~\citep{frantar2023gptq}, but under their authors' own setups and rarely with plain round-to-nearest (RTN). For example, at the layer scope chosen by QVLA~\citep{xu2026qvla}, our plain AWQ reaches 96.6\% on OpenVLA-OFT~\citep{kim2025oft}, against the 92.5\% QVLA reports for AWQ and the 96.7\% it reports for itself. RTN, which needs no calibration, matches AWQ and GPTQ on all four models we test: \pifive{}~\citep{black2025pi05}, GR00T~N1.5~\citep{nvidia2025gr00t}, UniVLA~\citep{bu2025univla} and OpenVLA-OFT (\cref{sec:accuracy,sec:audit}).

\textbf{Is task success really preserved?}
VLA quantization papers report their main results on standard LIBERO~\citep{liu2023libero}, with 50 (QuantVLA~\citep{zhang2026quantvla}), 250 (QVLA~\citep{xu2026qvla}) or 500 (ActQuant~\citep{akbari2026actquant}) episodes per suite and no perturbations. LIBERO is near saturation (93--98\% for the full-precision policies we test), and small samples hide small losses: at 50 episodes per suite, the 95\% interval of the four-suite average is $\pm$4.7 points. On the more recent paired LIBERO-plus tasks~\citep{fei2025liberoplus}, configurations with no significant LIBERO loss lose 1.5--4.3 points (\cref{sec:robust}).

\textbf{Is the deployed policy faster, and what actually runs?}
Three of the four releases we audit only simulate quantization in floating point, so their speed cannot be measured (\cref{sec:audit}). The bit width alone also says little about speed: on \pifive{}, the same 4-bit AWQ weights run $1.12$, $1.26$ or $6.8\times$ slower than F16, depending on which kernel executes them. Even ActQuant's real 3--4-bit models, which is faster than eager PyTorch, are $1.25$--$1.37\times$ slower than the same model in 16 bits on its own engine (\cref{sec:efficiency,sec:hardware}). Speed-ups must be measured against the same model on the same engine.

\begin{figure*}[t]
\centering
\includegraphics[width=\textwidth]{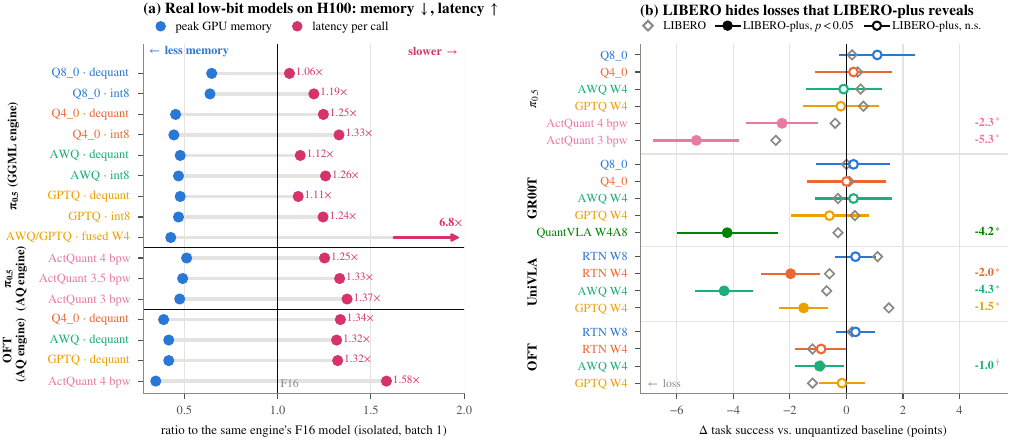}
\vspace{-2mm}
\caption{\textbf{The two headline results.} (a) Real low-bit \pifive{} and OpenVLA-OFT (OFT) on H100, relative to their 16-bit models: memory falls, but every configuration runs slower. (b) Change in task success: LIBERO (diamonds) shows almost no loss, while paired LIBERO-plus (circles, 95\% CI; filled: $p<0.05$) reveals losses for ActQuant, QuantVLA and 4-bit UniVLA. $^\dagger$Not significant after Holm correction.}
\label{fig:teaser}
\vspace{-2mm}
\end{figure*}

In this paper, we built \bench{} to answer these questions under one controlled protocol (\cref{sec:design}); concurrent work~\citep{xu2026vlaquantbench} studies layer scope and numerical format with RTN only. In every comparison only the quantizer changes: the checkpoint, the quantized layers and the engine stay the same, the kernel path and real or simulated execution are reported separately, and each quantized policy runs on exactly the same tasks as its unquantized baseline. 
% We evaluate RTN, AWQ, GPTQ and the VLA-specific ActQuant and QuantVLA on four VLAs (\pifive{}, GR00T~N1.5, UniVLA and OpenVLA-OFT) \textcolor{blue}{We need citations for each of these}, with 1{,}000--2{,}000 LIBERO~\citep{liu2023libero} episodes per configuration and thousands of paired LIBERO-plus~\citep{fei2025liberoplus} perturbation tasks (\cref{sec:protocol}). \pifive{} and OpenVLA-OFT run as real low-bit models; GR00T and UniVLA use simulated quantization at the scope of QuantVLA and QVLA (\cref{sec:models}). 
We evaluate RTN, AWQ~\citep{lin2024awq}, GPTQ~\citep{frantar2023gptq}, ActQuant~\citep{akbari2026actquant} and QuantVLA~\citep{zhang2026quantvla} on \pifive{}~\citep{black2025pi05}, GR00T~N1.5~\citep{nvidia2025gr00t}, UniVLA~\citep{bu2025univla} and OpenVLA-OFT~\citep{kim2025oft}, using 1{,}000--2{,}000 LIBERO~\citep{liu2023libero} episodes per configuration plus paired LIBERO-plus~\citep{fei2025liberoplus} tasks (\cref{sec:protocol}). \pifive{} and OpenVLA-OFT run real low-bit; GR00T and UniVLA use simulated quantization, as in QuantVLA and QVLA~\citep{xu2026qvla} (\cref{sec:models}). We time every policy in the closed loop, in isolation and per operator on an H100 NVL, repeat the efficiency study and a paired task-success subset on an NVIDIA L4 and a Jetson AGX Thor (\cref{sec:hardware}), and audit four public VLA-quantization releases (\cref{sec:audit}). \Cref{fig:teaser} shows the two headline results. To summarize, we make the following five contributions.

\begin{itemize}[leftmargin=1.2em]
\item \textbf{\bench{}.} We build the first VLA quantization benchmark \emph{that puts VLA-specific and generic methods at the same scope, times real low-bit models per kernel path on three GPUs, and tests robustness with paired distribution shift}. \footnote{\bench{} will be open-sourced upon acceptance.}
\item \textbf{Fewer bits, no speed-up.} We show why real low-bit quantized models need $2$--$3\times$ less memory, but can be significantly slower than 16 bit configurations ($1.06$--$1.58\times$ on library kernels) explaining why this occurs using Profiling. For example, the quantized backbones run one compute-bound pass, and \pifive{}'s memory-bound action expert, which takes 63\% of GPU time, stays in 16 bits. Even if the quantized layers took no time at all, a call would be at most $1.09\times$ faster (\cref{sec:anatomy,sec:efficiency}).
\item \textbf{LIBERO hides the losses.} PTQ configurations that look lossless on LIBERO lose up to 4.3 points under paired LIBERO-plus perturbations. Most of these losses come from the PTQ VLA-specific methods, ActQuant and QuantVLA, and from 4-bit UniVLA; plain RTN does as well as AWQ and GPTQ on all four models (\cref{sec:accuracy,sec:robust}).
\item \textbf{The GPU decides the speed, the method decides the success.} The same 4-bit files run up to 14\% faster on an L4 with int8 kernels but up to $1.9\times$ slower on a Jetson AGX Thor, while task-success outcomes agree on 90--99\% of episodes across the three GPUs (\cref{sec:hardware}).
\item \textbf{What a release lets you check.} Of four VLA quantization releases, only ActQuant can be timed, and it runs slower than its own 16-bit model. QVLA's reported margin over AWQ disappears once AWQ runs under our protocol (\cref{sec:audit,sec:discussion}).
\end{itemize}

\section{Background and Related Work}
\label{sec:related}
    \paragraph{VLA architectures.}
We study four VLAs that differ in how they generate actions. \textbf{\pifive{}}~\citep{black2025pi05} (3.36\,B parameters) encodes two camera images with SigLIP-So400m~\citep{zhai2023siglip}, runs a PaliGemma/Gemma-2B prefix~\citep{beyer2024paligemma} once per call to fill the KV cache, and generates a 10-step action chunk with a Gemma-300M-based \emph{action expert} (0.43\,B parameters) over 10 flow-matching steps. \textbf{GR00T~N1.5}~\citep{nvidia2025gr00t} pairs an Eagle-2.5 VLM with a diffusion-transformer action head (8 denoising steps). \textbf{UniVLA}~\citep{bu2025univla} is a 7\,B autoregressive VLA that predicts discrete latent-action tokens, while \textbf{OpenVLA-OFT}~\citep{kim2025oft} uses a similar Llama-2 backbone but predicts continuous action chunks with an MLP head.

\paragraph{PTQ and low-bit kernels for LLMs.}
Weight-only methods such as RTN, GPTQ~\citep{frantar2023gptq} and AWQ~\citep{lin2024awq} target memory-bound LLM decoding, where less weight traffic means lower latency. LLM.int8()~\citep{dettmers2022llmint8} brought 8-bit inference to scale mainly for memory, and weight-activation methods such as SmoothQuant~\citep{xiao2023smoothquant}, QuaRot~\citep{ashkboos2024quarot} and DuQuant~\citep{lin2024duquant} quantize both operands to use integer arithmetic. Actual speed-ups depend heavily on the kernel and batch size~\citep{zhao2024atom,lin2025qserve,frantar2025marlin}. VLAs differ from LLM decoding as the mostly process a prompt of several hundred tokens in one pass and then run a short action loop, so it is unclear which phase dominates and whether low-bit kernels help.

\paragraph{VLA-specific PTQ.}
ActQuant~\citep{akbari2026actquant} assigns mixed sub-4-bit llama.cpp formats using an action-guided sensitivity measure; its reported speed-up is measured on OpenVLA, comparing a 4-bit build in its C++ engine with PyTorch (\cref{sec:audit}). QuantVLA~\citep{zhang2026quantvla} builds on DuQuant-style W4A8 quantization with attention-temperature and head-balancing corrections, QVLA~\citep{xu2026qvla} assigns per-channel bit widths by action sensitivity, and HoloQ-VLA~\citep{wang2026holoq} targets uniform W4A4 with block-wise rotations. All four report their main results on standard LIBERO, but they quantize different layers, use different numbers of episodes, and run real or simulated kernels, so their results are hard to compare; none tests robustness under perturbation. Concurrent work~\citep{xu2026vlaquantbench} benchmarks layer scope and numerical format with RTN, but not VLA-specific methods, real low-bit speed, or robustness.

\paragraph{Robustness evaluation.}
LIBERO-plus~\citep{fei2025liberoplus} perturbs LIBERO tasks along seven axes (camera viewpoint, robot initial state, language, lighting, background, sensor noise and object layout), each at five difficulty levels, and shows that even full-precision VLAs struggle under these shifts. We ask a narrower question: does quantization \emph{add} failures beyond the base policy's? We run both policies on the same perturbed tasks and use paired tests, so that the base policy's own failures are not blamed on quantization.

\section{Benchmark Design}
\label{sec:design}
     \bench{} follows five design principles: (1) \textbf{Change only the quantizer.} Every variant of a model uses the same checkpoint, the same quantized layers, taken from the VLA-specific methods themselves, and the same calibration data (\cref{sec:models}); (2) \textbf{Report how the model runs, not just its bit width} (\cref{tab:taxonomy}). The same ``W4A16'' file can run in very different ways; in llama.cpp~\citep{llamacpp}, for example, it runs on int8 kernels that also quantize activations. We therefore report each kernel path separately; (3) \textbf{Use enough episodes, and pair them} (\cref{sec:protocol}). Each configuration runs 1{,}000--2{,}000 LIBERO episodes and is compared with its baseline task by task on thousands of LIBERO-plus tasks, so losses of one to two points become visible; (4) \textbf{Time the full policy on the same engine} (\cref{sec:measure}). The engine alone changes the latency of the same 16-bit \pifive{} by $2.1\times$, so every speed-up is measured against the unquantized model on the same engine; and (5) \textbf{Check what each release can actually reproduce} (\cref{sec:audit}): does it run, does it match its paper, and can its memory and speed be measured?

\subsection{Models, scopes and engines}
\label{sec:models}

\Cref{tab:models} lists the four VLAs, how each runs and how many episodes each configuration gets. To keep comparisons fair, every method on a model quantizes the same layers, taken from the VLA-specific method we compare against: ActQuant's vision and language backbone on \pifive{} and on real low-bit OpenVLA-OFT, QuantVLA's 116 layers (including the DiT action head's MLPs) on GR00T~N1.5, and QVLA's 435 backbone layers on UniVLA and simulated OpenVLA-OFT. \textbf{\pifive{}} asks what real low-bit execution delivers; we build a llama.cpp-based CUDA runtime with int8 MMQ, cuBLAS and fused W4A16 kernels, and also run ActQuant's engine and the official OpenPI server. \textbf{GR00T~N1.5} tests QuantVLA against generic methods inside QuantVLA's server. \textbf{UniVLA} and \textbf{OpenVLA-OFT} allow a comparison with QVLA's reported results and contrast discrete with continuous actions. For \textbf{real low-bit OpenVLA-OFT}, we regenerate ActQuant's 4- and 3-bit files with the authors' settings, since no checkpoints are released.

AWQ uses the official llm-awq search and clipping; GPTQ uses our implementation, verified against the reference code (99.7--99.997\% identical codes on identical Hessians). Both use the same 512 calibration frames per model, drawn from ActQuant's 60-episode calibration set for \pifive{} and GR00T (per suite for GR00T), and from 512 LIBERO training trajectories for UniVLA and OpenVLA-OFT, following QVLA.

\begin{table}[!htbp]
    \centering
    \caption{\textbf{Models and evaluation size.} Configurations include the unquantized baseline. \emph{Real} models run true low-bit weights and are timed; \emph{sim.} models simulate quantization and are tested for task success only. The last two columns give episodes per configuration; on LIBERO-plus, every configuration runs the same tasks.}
    \label{tab:models}
    \scriptsize
    \setlength{\tabcolsep}{3pt}
    % Quantized-scope column commented out; restore "l" in the column spec and the cells below to bring it back.
    \begin{tabular}{@{}l l c c c@{}}
    \toprule
    Model & % Quantized scope &
    Execution & Configurations & LIBERO & LIBERO-plus\\
    \midrule
    \pifive{} & % SigLIP + PaliGemma + emb.\ (87\%) &
    real & 18 & 2{,}000 & 2{,}016$^a$\\
    GR00T N1.5 & % QuantVLA's 116 linears (LLM + DiT MLPs) &
    sim. & 6 & 2{,}000 & 2{,}016\\
    UniVLA & % QVLA's 435 modules &
    sim. & 5 & 1{,}000 & 5{,}014\\
    OpenVLA-OFT & % QVLA's 435 modules &
    sim. & 5 & 1{,}000 & 5{,}014\\
    OpenVLA-OFT & % ActQuant's 532 tensors &
    real & 10$^b$ & 2{,}000 & 2{,}016\\
    \bottomrule
    \end{tabular}\\[2pt]
    % {\scriptsize $^a$10 configurations plus one diagnostic: GGML F16, and RTN Q8\_0/Q4\_0, AWQ and GPTQ on the cuBLAS path; ActQuant-engine F16, ActQuant 4 and 3\,bpw, and RTN Q3\_K (W3A8); PyTorch BF16; plus one fixed-noise F16 diagnostic. $^b$Efficiency: 10 configurations on the L4 and Jetson AGX Thor, 6 on the H100; task success: 5 configurations on the cross-GPU subset (\cref{sec:hardware}).}
    \scriptsize $^a$10 of the 18 configurations, plus one fixed-noise diagnostic (\cref{tab:pi05}). $^b$Task success for 7 of the 10 configurations (\cref{tab:robust}); the others are efficiency-only. A 5-configuration subset also runs on all three GPUs (\cref{sec:hardware}).

\end{table}

\subsection{Evaluation protocol}
\label{sec:protocol}
We use two robotic benchmarks: First, we use LIBERO's four suites, each with 10 tasks and 50 trials (2{,}000 episodes) per configuration. We use the full suite with \pifive{} and GR00T, but limit the trial to 25 trials (1{,}000 episodes) when evaluating UniVLA and OpenVLA-OFT, to be comparable to QVLA. All runs use fixed seeds and official step limits.
The second benchmark we use is LIBERO-plus. In our experiments, every quantized policy runs on exactly the same perturbed tasks as its baseline: 72 per category and suite for \pifive{} and GR00T (2{,}016 tasks), and a fixed 50\% subset stratified by suite and category for UniVLA and OpenVLA-OFT (5{,}014 tasks). Absolute scores are comparable only within each pair.

% \textbf{Statistics.} \textcolor{blue}{not clear what we are saying here}Success rates carry Wilson 95\% intervals~\citep{wilson1927}; LIBERO differences use unpaired Newcombe intervals~\citep{newcombe1998}, and LIBERO-plus differences an exact paired McNemar test~\citep{mcnemar1947} that treats sampling noise as part of each outcome. Effects we call significant survive Holm correction~\citep{holm1979} unless marked \textcolor{blue}{what is this?}$^\dagger$; noise diagnostics and per-category tests are exploratory.

\textbf{GPUs.} We run all benchmarking experiments on an H100 GPU cluster. In addition, we run a scaled down set of experiments on the L4 and Jetson AGX Thor GPUs due to time and cost limitations. In the scaled-down experiments, the same byte-identical files run all efficiency measurements, and task success uses a paired subset of 15 configurations, 120 LIBERO episodes and 252 LIBERO-plus tasks drawn from the H100 runs (\cref{sec:hardware}).
In total, \bench{} comprises 170{,}524 closed-loop episodes on the H100 and 11{,}310 on the L4 and Thor.

\subsection{Efficiency measurement}
\label{sec:measure}
We measure efficiency at three levels. (1) \textbf{Isolated} (all four VLAs): one configuration at a time on an idle GPU, 10 warm-up and 50 measured calls, with per-process peak memory and GPU energy from the hardware counter; these runs give our exact ratios. (2) \textbf{Closed loop} (\pifive{}): the median latency of $\approx$65{,}000 calls over 2{,}000 LIBERO episodes; since it varies by up to 24\% across days and GPUs, we use it only to confirm the direction of effects. (3)\textbf{Operator level}: Nsight Systems on the same six LIBERO frames, giving GPU time per operator and component, host overhead, and a roofline class for each weight multiplication~\citep{williams2009roofline}. 
% (H100 ridge: 214\,FLOP/B).

\section{Evaluating PTQ for VLAs}
\label{sec:anatomy}
    \subsection{Can weight-only quantization make a VLA faster?}
Weight-only PTQ (AWQ, GPTQ) speeds up LLM decoding, where each step processes one token and loading weights dominates. Most VLA quantization methods carry this idea over and compress the vision-language backbone. We test whether that pays off by profiling the official unquantized \pifive{} (OpenPI BF16, batch 1, 10 flow steps, H100 NVL). A matrix multiplication is \emph{memory-bound} when weight loading dominates, so fewer weight bits help, and \emph{compute-bound} when arithmetic dominates, so they do not; the boundary, the \emph{ridge}, is $\approx$214\,FLOP/B on the H100 (roofline in \cref{fig:roofline}).

\begin{figure}[t]
    \centering
    \includegraphics[width=\columnwidth]{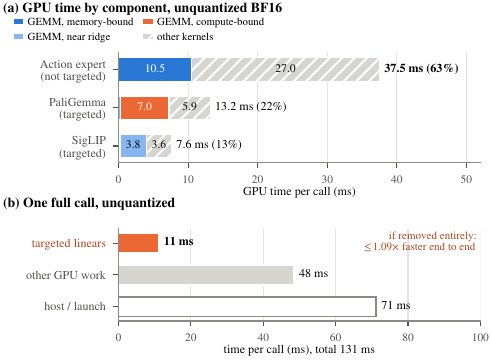}
    \vspace{-2mm}
    \caption{\textbf{Quantization targets only a small part of \pifive{}'s time} (unquantized BF16, batch 1, H100). (a) The untargeted action expert takes 63\% of GPU time. (b) The targeted layers take 11 of 131\,ms per call: at most $1.09\times$ speed-up.}
    \label{fig:anatomy}
    \vspace{-3mm}
\end{figure}

\textbf{The quantized layers are not where the time goes.}
Of a 130.5\,ms call, 59.4\,ms is GPU work and 71.1\,ms is host and launch overhead (\cref{fig:anatomy}). The quantized layers are not memory-bound: PaliGemma processes the whole 968-token prompt in one pass, like an LLM prefill, and is compute-bound (median 498\,FLOP/B, 61\% of FP16 peak); SigLIP sits just below the ridge (177\,FLOP/B). The memory-bound part is the action expert: it processes only 10 tokens per flow step ($\approx$10\,FLOP/B), and with 12.8\% of the parameters it takes 63\% of GPU time. Of its 37.5\,ms, only 10.5\,ms are weight multiplications, the part weight-only quantization can speed up; the other 27.0\,ms are attention, norms and element-wise kernels. Every method we run on \pifive{}, including ActQuant, keeps the action expert in 16 bits.

Even if the quantized layers (11.0\,ms) took no time, GPU time would improve by at most $59.4/48.4=1.23\times$ and the full call by $130.5/119.5=1.09\times$. Quantizing the action expert's weights instead would save at most its 10.5\,ms of weight multiplications.
When evaluating with different engines, the picture does not change (roofline in \cref{app:roofline}). In our GGML engine the three components fall into the same regimes, only less efficiently (PaliGemma at 22\% vs.\ 61\% of peak; Appendix\ref{app:roofline}).

Profiling the other three VLAs the same way (\cref{tab:cross_e2}) shows that no single layer set is the right target. GR00T, like \pifive{}, spends most GPU time in its action head (58\%, mostly memory-bound). OpenVLA-OFT spends 76\% in its LLM, but in one parallel pass over $\approx$600 tokens, which is compute-bound. Only UniVLA, which decodes tokens one at a time, spends its time (88\%) in memory-bound or near-ridge LLM layers, the case weight-only quantization targets.

\takeaway{Weight-only PTQ speeds up LLM decoding, but \pifive{}'s backbone runs one compute-bound pass, so compressing it can speed up a call by at most $1.09\times$, while the memory-bound action expert (63\% of GPU time) stays in 16 bits. Profiling before quantizing can thus explain the right quantization target depending on the model.}

\section{Accuracy on LIBERO}
\label{sec:accuracy}
    We now focus on the accuracy effects of PTQ for VLAs compared to generic PTQ methods. We only evaluate model the original authors' open-sourced their checkpoints for to make sure that we do not introduce errors when evaluating those papers with VLAs that are not supported or open-sourced by the authors.  \Cref{tab:robust} summarizes LIBERO success for all four VLAs; all 18 \pifive{} configurations, with per-suite success, latency and memory, are in \cref{tab:pi05} (\cref{app:libero}).

\textbf{On LIBERO, 8- and 4-bit methods look almost the same.}
On \pifive{}, every RTN, AWQ and GPTQ configuration reaches 96.7--97.6\%, within or above the interval of the unquantized F16 (FP16) model (96.5\%, 95.6--97.2), on every kernel path except AWQ on MMQ, which fails numerically (see below). With 2{,}000 episodes per configuration, any loss above $\approx$1.5 points would be detected; none is. ActQuant at 4\,bpw also matches its engine's F16 (96.6 vs.\ 97.0\%, n.s.); on real low-bit OpenVLA-OFT it loses a small but significant 1.3 points (95.4 vs.\ 96.6\%), while RTN, AWQ and GPTQ stay within 0.6. The other models agree: on GR00T all five quantized configurations, QuantVLA included, are within 0.3 points of FP; UniVLA scores 92.0--94.2\% against 92.7\%; simulated OpenVLA-OFT 96.3--97.7\% against 97.5\%, where plain AWQ (96.6\%) matches QVLA's reported 96.7\%. Calibration brings no measurable benefit: plain RTN matches AWQ and GPTQ on all four models. This is consistent with LLMs: on LLaMA models, AWQ and GPTQ improve on RTN mainly at 3 bits, and their perplexity gain at 4 bits with group size 128 is small~\citep{lin2024awq}; in closed-loop task success, this small gain is no longer visible. Standard LIBERO therefore shows no accuracy difference between generic (calibrated or not) and VLA-specific PTQ at 4 bits and above, apart from a 1.3-point ActQuant loss on OpenVLA-OFT. The only large gap is AWQ on MMQ, which drops to 84.4\% on the \texttt{Goal} suite: on one frame, an activation in PaliGemma's prefix exceeds the FP16 maximum, and the overflow crashes the server, so the remaining episodes count as failures. On the 432 episodes that run, AWQ reaches 97.7\%. Replaying that frame fails on every AWQ kernel but on no other method. We believe AWQ's per-channel scaling is the cause: it divides activations by a per-channel factor, which enlarges channels whose factor is below one. AWQ's offline action error (3.6\%) gives no warning of this.
% ActQuant is \textcolor{blue}{the only VLA specific technique that supports 3 bit quantization, On LIBERO,} 
% against its own engine's F16, ActQuant loses 

\textbf{At 3 bits, ActQuant is no better than plain rounding.}
Of the VLA-specific methods we run, only ActQuant offers 3-bit models. On \pifive{}'s LIBERO, against its own engine's F16, it loses 0.4 points at 4\,bpw and 0.9 at 3.5\,bpw (both n.s.), but 2.5 at 3\,bpw, mostly on the long-horizon suite (87.0 vs.\ 93.8\%), where small action errors accumulate over up to 520 steps. 
% Plain RTN does just as well \textcolor{blue}{so it is better or worse?} on the same engine and int8-activation path: at 3 bits (Q3\_K, slightly larger at 2.47 vs.\ 2.36\,GB) it reaches 95.7\% on LIBERO and is statistically indistinguishable from ActQuant on LIBERO-plus (+0.5 points, $p=0.56$). 
Plain RTN on the same engine and int8-activation path is no worse: at 3 bits (Q3\_K, slightly larger at 2.47 vs.\ 2.36\,GB) it reaches 95.7\% on LIBERO, 1.2 points above ActQuant (n.s.), and is statistically indistinguishable from it on LIBERO-plus (+0.5 points, $p=0.56$).
We use RTN as this matched baseline because it is the weakest generic method: if plain rounding matches ActQuant, ActQuant's sensitivity-guided mix of bit widths brings no measurable benefit. 

\textbf{Offline action error does not predict success.}
On \pifive{}, action error on held-out frames varies about 20-fold across RTN, AWQ and GPTQ (0.2\% for Q8\_0 to 3.6\% for AWQ), yet closed-loop success is the same (\cref{app:fidelity}). The reverse also holds: ActQuant at 4\,bpw has a similar error to AWQ (3.3 vs.\ 3.6\%) but loses 2.3 points on LIBERO-plus, where AWQ loses nothing. On GR00T, QuantVLA's error (15\%) is 9 to 30 times that of 4-bit RTN, AWQ and GPTQ, yet its LIBERO success is unchanged. Offline error also gave no warning of AWQ's FP16 overflow, so it cannot replace closed-loop evaluation.

\takeaway{On LIBERO, every 8- and 4-bit method, generic or VLA-specific, calibrated or not, matches full precision; only ActQuant on OpenVLA-OFT (1.3 points), the 3-bit models on the int8-activation path and AWQ's FP16 crash lose. LIBERO is too saturated to rank methods, and offline action error cannot predict failures: differences appear only under distribution shift (\cref{sec:robust}).}

\section{There is no free lunch: Memory reductions and Latency}
\label{sec:efficiency}
    \begin{table}[!htbp]
\centering
\caption{\textbf{Isolated efficiency on the H100: quantization cuts memory but raises latency and energy.} Batch 1, 50 calls after warm-up; $\times$base: latency vs.\ the same engine's F16. $^{\dagger}$Later session. $^{\ddagger}$Separate session with its own ActQuant-engine F16 (88.5\,ms; ActQuant 3\,bpw: 119.3\,ms, $1.35\times$); ratio vs.\ that F16.}

\label{tab:isolated}
\scriptsize
\setlength{\tabcolsep}{2.4pt}
\resizebox{\columnwidth}{!}{%
\begin{tabular}{@{}l c c c c c@{}}
\toprule
Configuration & File GB & Peak GiB & ms/call & $\times$base & J/call\\
\midrule
\grouprow{6}{\pifive{}, GGML engine}
F16 & 6.71 & 7.04 & 66.8 & 1.00 & 13.8\\
Q8\_0 cuBLAS / MMQ & 3.97 & 4.55 / 4.48 & 71.0 / 79.7 & 1.06 / 1.19 & 15.5 / 16.6\\
Q4\_0 cuBLAS / MMQ$^\dagger$ & 2.51 & 3.19 / 3.12 & 83.1 / 88.7 & 1.24 / 1.33 & 16.5 / 17.5\\
AWQ cuBLAS / MMQ & 2.70 & 3.36 / 3.30 & 74.9 / 83.9 & 1.12 / 1.26 & 15.9 / 16.9\\
GPTQ cuBLAS / MMQ & 2.70 & 3.36 / 3.30 & 74.1 / 83.0 & 1.11 / 1.24 & 15.5 / 16.8\\
AWQ / GPTQ fused W4 & 2.40 & 3.01 / 3.00 & 453 / 453 & 6.79 / 6.78 & 118 / 118\\
\grouprow{6}{\pifive{}, ActQuant engine}
F16 (our GGUF) & 6.71 & 6.73 & 84.9 & 1.00 & 20.9\\
ActQuant 4 / 3.5 / 3 bpw & 2.63--2.36 & 3.45--3.20 & 106--117 & 1.25--1.37 & 24.8--26.1\\
RTN Q3\_K (W3A8-dyn)$^\ddagger$ & 2.47 & 3.30 & 115.4 & 1.30 & 26.5\\
\grouprow{6}{\pifive{}, OpenPI PyTorch}
BF16 eager / compiled & 7.23 & 7.94 / 8.03 & 124.2 / 32.3 & -- & 19.7 / 9.6\\
\grouprow{6}{GR00T N1.5, PyTorch}
FP (BF16) & 7.59 & 6.08 & 124.6 & 1.00 & 17.3\\
QuantVLA (simulated W4A8) & 7.59 & 10.46 & 237.2 & 1.90 & 29.8\\
\bottomrule
\end{tabular}}
\vspace{-3mm}
\end{table}

We now focus on the memory and storage reductions of PTQ methods, focusing on how these reductions also affect the latency. Table~\ref{tab:isolated} shows both File size reductions and peak memory reductions for the \pifive{} and GR00T. At 4 bits real quantization, \pifive{} shrinks the file sizes by up to $2.8\times$ (6.71 to 2.40--2.70\,GB) and the isolated peak memory by up to $2.3\times$ (7.04 to 3.0--3.5\,GiB; \cref{tab:isolated}); real low-bit OpenVLA-OFT falls from 15.6 to 5.4--6.5\,GiB on the H100 (\cref{tab:cross_energy}). The gain stays well below $4\times$ because the action expert, projector, activations, KV cache and workspace remain in 16 bits. We note that the compact 2.40\,GB W4-g128 file runs only on our slow fused kernel (library kernels need the 2.70\,GB Q4\_1 re-encoding), and that simulated QuantVLA \emph{grows} memory by $1.7\times$.

\textbf{No quantized configuration is faster.}
% All 14 quantized \pifive{} configurations are slower than their engine's F16, in isolation ($1.06$--$1.37\times$ on library kernels, $6.8\times$ fused) and in the closed loop ($1.10$--$1.40\times$, fused $5.4\times$; \cref{tab:pi05}), and so is real low-bit OpenVLA-OFT ($1.32$--$1.58\times$). Because GPU power barely changes (200--260\,W), energy per call rises with latency ($1.1$--$1.3\times$, fused $8.6\times$). For ActQuant, \emph{fewer bits are slower}: 106, 113 and 117\,ms at 4, 3.5 and 3\,bpw, as denser formats need more decoding work. Simulated QuantVLA on GR00T is $1.9\times$ slower than FP, since it keeps floating-point weights and adds rounding.
All 14 quantized \pifive{} configurations are slower than their engine's F16, in isolation ($1.06$--$1.37\times$ on library kernels, $6.8\times$ fused) and in the closed loop ($1.10$--$1.40\times$; \cref{tab:pi05}), and so is real low-bit OpenVLA-OFT ($1.32$--$1.58\times$). Since GPU power barely changes, energy per call rises with latency ($1.1$--$1.3\times$). Simulated QuantVLA is $1.9\times$ slower than FP on GR00T, because it keeps floating-point weights and adds rounding.

\begin{figure}[t]
\centering
\includegraphics[width=\columnwidth]{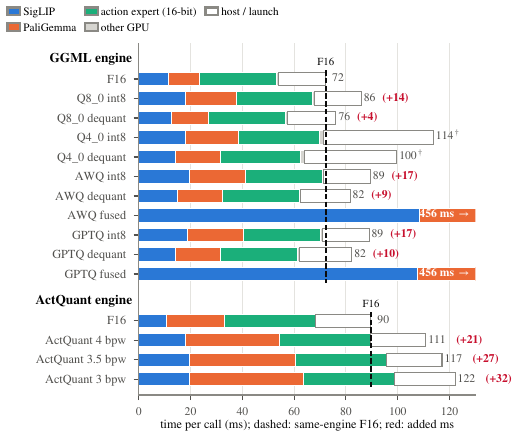}
\vspace{-2mm}
\caption{\textbf{Quantization adds time only where it is applied} (\pifive{}, Nsight, batch 1, same six frames). SigLIP and PaliGemma grow on every quantized path; the 16-bit action expert stays constant. $^\dagger$Q4\_0 was profiled in a separate session with higher host overhead: its GPU times are comparable, its totals are not.}
\label{fig:components}
\vspace{-3mm}
\end{figure}

\textbf{Every quantized layer is slower, on identical shapes.}
Comparing exactly the quantized layers with the same F16 layers on the same engine (\cref{fig:components,tab:slowdown}), \emph{not one} of the 283--288 quantized operators in any of the 13 configurations is faster: they take $1.24\times$ (Q8\_0, cuBLAS) to $2.44\times$ (ActQuant 3\,bpw) the F16 time, and $28.7\times$ on the fused kernel. The compute-bound PaliGemma slows down most on every path, while the 16-bit action expert stays constant (29--32\,ms on GGML, 35\,ms on ActQuant's engine), confirming that the extra time comes only from quantization.
Whle quantization saves memory traffic, it adds decoding work. That trade-off only pays for memory-bound layers, and \pifive{}'s quantized layers are compute-bound or near the ridge (\cref{fig:roofline}). On the H100,  every kernel path is slower as dequantizing rebuilds FP16 weights on every call. In addition, int8 MMQ also quantizes activations and runs slower kernels than cuBLAS, and the fused W4A16 kernel is far less efficient than cuBLAS. On the roofline, the quantized layers use fewer bytes but run at lower TFLOP/s (\cref{fig:qroofline}).

\textbf{Speed depends on the engine as much as on the quantizer.}
% Before quantizing anything, the same 16-bit \pifive{} already takes 47\,ms per closed-loop call in compiled PyTorch, 81\,ms in our GGML engine and 99\,ms in ActQuant's engine. This $2.1\times$ spread is larger than the effect of any quantizer we measured on library kernels, although faster weight-activation kernels could change this. A quantization speed-up is therefore meaningful only against the same model on the same engine; ActQuant's reported speed-up, by contrast, compares its C++ build with PyTorch (\cref{sec:audit}).
Before quantizing anything, the same 16-bit \pifive{} takes 47\,ms per closed-loop call in compiled PyTorch, 81\,ms in our GGML engine and 99\,ms in ActQuant's engine. This $2.1\times$ spread is larger than any quantizer's effect on library kernels, so speed-ups must be measured on the same engine; ActQuant's reported speed-up compares its C++ build with PyTorch (\cref{sec:audit}).

\cref{fig:components} shows that all of the added time lands in SigLIP and PaliGemma. On ActQuant's engine, PaliGemma alone grows from 22.2\,ms in F16 to 43.9\,ms at 3\,bpw, which is why fewer bits run slower. Host overhead (18--20\,ms) is already larger than the entire F16 time of the quantized layers ($\approx$14\,ms). Memory has a 16-bit floor too: dropping ActQuant from 4 to 3\,bpw saves only 0.25\,GiB of peak memory but costs another 3.0 points on LIBERO-plus.

\takeaway{Quantization makes \pifive{} smaller, not faster. On the H100 at batch 1, 4-bit weights cut peak memory by up to $2.3\times$ (8-bit: $1.55\times$), yet every quantized layer runs slower, so the full policy takes 6--37\% longer and 12--27\% more energy per call on library kernels (OpenVLA-OFT: 32--58\% longer). Fewer bits even run slower, and the choice of engine changes latency more than any quantizer: speed claims need same-engine, end-to-end measurements.}

\section{Robustness Under Distribution Shift}
\label{sec:robust}
    \begin{table}[!htbp]
\centering
\caption{\textbf{LIBERO hides losses that LIBERO-plus reveals.} Success rate (\%) and its change $\Delta$ (points) against the unquantized baseline, the first row of each block (``--''). The baseline runs on the same engine and, on LIBERO-plus, on the same tasks. \textbf{Bold}: significant loss ($p<0.05$); $^\dagger$not significant after correcting for multiple tests. \pifive{} and OFT (real) run real low-bit models; GR00T, UniVLA and OFT (sim.) simulate quantization. UniVLA and OFT (sim.) use a different LIBERO-plus task sample.}
\label{tab:robust}
\scriptsize
\setlength{\tabcolsep}{2.0pt}
\scalebox{0.9}{%
\begin{tabular}{@{}l l r r r r@{}}
\toprule
 & & \multicolumn{2}{c}{LIBERO} & \multicolumn{2}{c}{LIBERO-plus}\\
\cmidrule(lr){3-4}\cmidrule(lr){5-6}
Model & Config & \% & $\Delta$ & \% & $\Delta$\\
\midrule
\multirow{9}{*}{\pifive{}} & GGML F16 & 96.5 & -- & 86.6 & --\\
 & RTN Q8\_0 & 96.7 & +0.2 & 87.7 & +1.1\\
 & RTN Q4\_0 & 96.9 & +0.4 & 86.9 & +0.2\\
 & AWQ W4 & 97.0 & +0.5 & 86.5 & $-$0.1\\
 & GPTQ W4 & 97.1 & +0.6 & 86.4 & $-$0.2\\
 & AQ-engine F16 & 97.0 & -- & 82.9 & --\\
 & ActQuant 4 bpw & 96.6 & $-$0.4 & 80.7 & \textbf{$-$2.3}\\
 & ActQuant 3 bpw & 94.5 & \textbf{$-$2.5} & 77.6 & \textbf{$-$5.3}\\
 & RTN Q3\_K (W3A8) & 95.7 & \textbf{$-$1.3} & 78.1 & \textbf{$-$4.8}\\
\midrule
\multirow{7}{*}{\shortstack[l]{OFT\\(real)}} & AQ-engine F16 & 96.6 & -- & 73.0 & --\\
 & RTN Q8\_0 & 96.9 & +0.3 & 74.0 & +1.0\\
 & RTN Q4\_0 & 96.7 & +0.1 & 72.3 & $-$0.6\\
 & AWQ W4 & 96.4 & $-$0.3 & 71.7 & $-$1.3\\
 & GPTQ W4 & 97.2 & +0.6 & 73.1 & +0.1\\
 & ActQuant 4 bpw & 95.4 & \textbf{$-$1.3} & 68.1 & \textbf{$-$4.9}\\
 & ActQuant 3 bpw & 94.9 & \textbf{$-$1.7} & 64.5 & \textbf{$-$8.4}\\
\midrule
\multirow{6}{*}{GR00T} & FP & 93.0 & -- & 78.2 & --\\
 & QuantVLA W4A8 & 92.8 & $-$0.3 & 74.0 & \textbf{$-$4.2}\\
 & RTN Q8\_0 & 93.0 & 0.0 & 78.5 & +0.2\\
 & RTN Q4\_0 & 93.1 & +0.1 & 78.2 & 0.0\\
 & AWQ W4 & 92.7 & $-$0.3 & 78.5 & +0.2\\
 & GPTQ W4 & 93.3 & +0.3 & 77.6 & $-$0.6\\
\midrule
\multirow{5}{*}{UniVLA} & FP & 92.7 & -- & 52.5 & --\\
 & RTN W8 & 93.8 & +1.1 & 52.8 & +0.3\\
 & RTN W4 & 92.1 & $-$0.6 & 50.5 & \textbf{$-$2.0}\\
 & AWQ W4 & 92.0 & $-$0.7 & 48.2 & \textbf{$-$4.3}\\
 & GPTQ W4 & 94.2 & +1.5 & 51.0 & \textbf{$-$1.5}\\
\midrule
\multirow{5}{*}{\shortstack[l]{OFT\\(sim.)}} & FP & 97.5 & -- & 73.1 & --\\
 & RTN W8 & 97.7 & +0.2 & 73.4 & +0.3\\
 & RTN W4 & 96.3 & $-$1.2 & 72.2 & $-$0.9\\
 & AWQ W4 & 96.6 & $-$0.9 & 72.1 & \textbf{$-$1.0}$^\dagger$\\
 & GPTQ W4 & 96.3 & $-$1.2 & 72.9 & $-$0.2\\
\bottomrule
\end{tabular}}
\vspace{-3mm}
\end{table}

% Every policy, quantized or not, loses success under perturbation: 9--17 points on \pifive{}, 14--19 on GR00T, about 24 on OpenVLA-OFT and 40--44 on UniVLA. That drop belongs to the base policy. Our question is narrower: does quantization \emph{add} failures, and does LIBERO see them? Because each quantized policy runs on exactly the same perturbed tasks as its baseline, a paired test answers this directly (\cref{tab:robust}, \cref{fig:teaser}b).
We now shift our focus on the robustness of the PTQ methods. We focus our experiments on how running with LIBERO-plus affects the results as shown in Table~\ref{tab:robust}. Every policy we evaluate has reduced success under perturbation compared to the LIBERO results (9--17 points on \pifive{}, 40--44 on UniVLA). Most of the drop is due to the base VLA itself showing less success with LIBERO-plus. We ask whether quantization \emph{adds} failures, and whether LIBERO sees them, by running each quantized policy on exactly the same perturbed tasks as its baseline (\cref{tab:robust}, \cref{fig:teaser}b).

On \pifive{} and GR00T, RTN, AWQ and GPTQ at 8 and 4 bits add no measurable loss: all eight comparisons stay within $\pm$1.1 points compared to the baseline.  On UniVLA and OpenVLA-OFT, 8-bit weights are just as safe (+0.3 on both), and on OpenVLA-OFT, real or simulated, even 4-bit weights stay within 1.3 points.

However, VLA-specific methods show noticeable reduction of performance between LIBERO and LIBERO-plus.
On GR00T, where every method quantizes the same 116 layers chosen by QuantVLA, including the DiT action head's MLPs, generic W4 methods lose at most 0.6 points, but QuantVLA loses 4.2, and 12.3 on the \texttt{Goal} suite alone, while its LIBERO score is unchanged. On \pifive{}, ActQuant loses 2.3 points at 4\,bpw and 5.3 at 3\,bpw against its engine's F16, where LIBERO showed only $-0.4$ and $-2.5$. Vanilla RTN at the same 3 bits and int8-activation path loses about as much ($-4.8$; $+0.5$ against ActQuant). On OpenVLA-OFT, where F16 and every method share ActQuant's engine and layers, ActQuant loses 4.9 points at 4\,bpw and 8.4 at 3\,bpw, while RTN, AWQ and GPTQ stay within 1.3 (n.s.). 
% Both VLA-specific methods also quantize activations, a plausible but unproven cause, since QuantVLA also changes the weight quantizer. Either way, at 4 bits and the same scope the generic methods are more robust; at 3 bits, RTN and ActQuant lose alike.

\begin{figure}[t]
    \centering
    \includegraphics[width=\columnwidth]{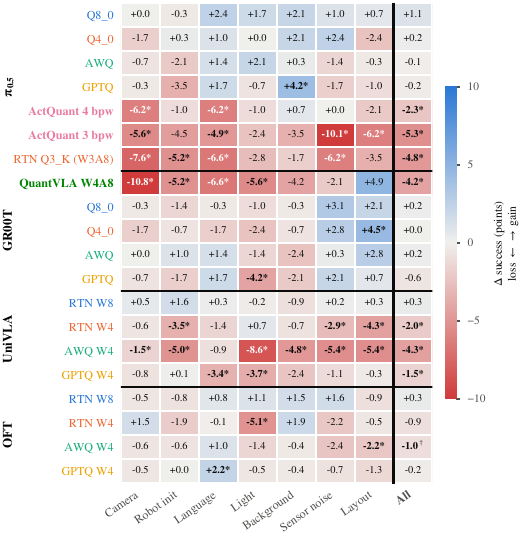}
    \vspace{-2mm}
    \caption{\textbf{Where quantization loses under perturbation.} Paired change in LIBERO-plus success per perturbation category (points vs.\ the same-engine unquantized model); All: overall change. Losses concentrate in the VLA-specific methods (bold), in plain RTN at 3 bits on the same int8-activation path as ActQuant, and in 4-bit UniVLA. $^\ast$McNemar $p<0.05$, uncorrected; $^\dagger$not significant after Holm correction.}

    \label{fig:heatmap}
    \vspace{-3mm}
\end{figure}

% \textbf{Generic 4-bit weights are not safe everywhere.}
% The VLA-specific methods lose most under camera-viewpoint and language shifts (QuantVLA $-10.8$ and $-6.6$, ActQuant 4\,bpw $-6.2$ on both), and at 3\,bpw also under sensor noise and object layout ($-10.1$, $-6.2$; \cref{fig:heatmap}). Among generic methods, only 4-bit UniVLA loses broadly (AWQ $-8.6$ under lighting); on \pifive{} and GR00T just 3 of 56 cells are significant, about what chance predicts (\cref{app:groot}).
The VLA-specific methods lose most under camera-viewpoint and language shifts as shown in Figure\cref{fig:heatmap}, (QuantVLA $-10.8$ and $-6.6$, ActQuant 4\,bpw $-6.2$ on both), and at 3\,bpw also under sensor noise and object layout ($-10.1$, $-6.2$). Plain RTN at 3 bits loses in a similar way. Among generic 4-bit methods, only UniVLA loses broadly (AWQ $-8.6$ under lighting); on \pifive{} and GR00T just 3 of 56 cells are significant, about what chance predicts (\cref{app:groot}).

% \textbf{Which perturbations cause failures.}
% % The VLA-specific methods fail on the shifts a robot meets most: camera viewpoint and language (QuantVLA $-10.8$ and $-6.6$, ActQuant 4\,bpw $-6.2$ on both), and at 3\,bpw sensor noise and object layout ($-10.1$, $-6.2$; \cref{fig:heatmap}). Among generic methods, only 4-bit UniVLA shows broad losses (AWQ $-8.6$ under lighting); on OpenVLA-OFT the one sizeable drop is lighting under RTN W4 ($-5.1$), and on \pifive{} and GR00T just 3 of 56 cells are significant, about what chance predicts (\cref{app:groot}).
% The VLA-specific methods lose most under camera-viewpoint and language shifts (QuantVLA $-10.8$ and $-6.6$, ActQuant 4\,bpw $-6.2$ on both), and at 3\,bpw also under sensor noise and object layout (\cref{fig:heatmap}). Among generic methods, only 4-bit UniVLA loses broadly (AWQ $-8.6$ under lighting); on \pifive{} and GR00T just 3 of 56 cells are significant, about what chance predicts (\cref{app:groot}).

\begin{figure}[t]
    \centering
    \includegraphics[width=\columnwidth]{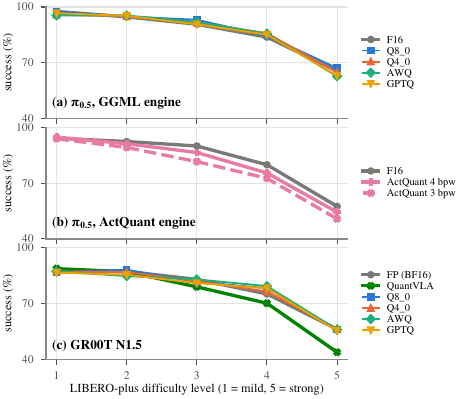}
    \vspace{-2mm}
    \caption{\textbf{LIBERO-plus success by perturbation difficulty.} Weight-only methods track their baselines at every level; ActQuant and QuantVLA fall behind from level 3 on, where perturbations are strongest. ActQuant is compared with F16 on its own engine, which scores lower than ours on LIBERO-plus for reasons unrelated to quantization.}
    \label{fig:difficulty}
    \vspace{-3mm}
\end{figure}

\cref{fig:difficulty} shows how losses grow with perturbation strength.
On mild perturbations (level 1), every \pifive{} and GR00T method stays within about two points of its baseline. From level 3 on, the VLA-specific methods fall behind: ActQuant at 3\,bpw by 6.7--8.4 points, and QuantVLA by up to 12.4 (43.8 vs.\ 56.2\% at level 5). Generic methods on OpenVLA-OFT stay within about two points throughout, while UniVLA's W4 losses appear already at level 1 (up to $-7.0$). Quantization error and distribution shift seem to compound. An error that is harmless on easy inputs can flip a decision the policy is already unsure about, and unperturbed LIBERO never tests such decisions (as discussed more in \cref{app:difficulty}).

\textbf{Calibration does not buy robustness here.}
Tested directly against plain RTN on the same LIBERO-plus tasks (\cref{app:calib}), neither AWQ nor GPTQ is significantly more robust on any of the four models. AWQ is even significantly worse on UniVLA ($-2.4$ points vs.\ RTN). Lower reconstruction error does not help either: on GR00T, GPTQ's action error (0.5\%) is a third of RTN's (1.6\%), yet it is not more robust ($-0.6$ vs.\ $0.0$).

\textbf{The engine alone shifts robustness, so it must be held fixed.}
The \emph{same} F16 file loses 3.7 points on LIBERO-plus when run on ActQuant's engine instead of ours, although both score the same on LIBERO. Our engine matches the official PyTorch policy (86.6 vs.\ 86.7\%). About half of the gap is consistent with ActQuant reusing one fixed noise sample per call, and the rest with differences in input preprocessing; both diagnostics are exploratory (\cref{app:noise}). ActQuant's own comparisons are unaffected, since both sides share its engine, but a comparison across engines would have charged the whole gap to the quantizer.

\takeaway{Under distribution shift, generic 8-bit weights are safe on all four VLAs, and generic 4-bit weights on the three with continuous actions; calibration adds no robustness over plain RTN. At 4 bits, the VLA-specific methods lose 2.3--4.9 points (at 3 bits, ActQuant 5.3--8.4 and RTN 4.8) and 4-bit UniVLA 1.5--4.3, losses that LIBERO misses or underestimates. Robustness claims need paired tests under distribution shift, on a fixed engine.}

\section{The effect of Hardware choice}
\label{sec:hardware}
    % Cross-hardware replication: NVIDIA L4 (AWS g6.2xlarge) and NVIDIA Jetson AGX Thor.
% Data: benchmark-Results/final-g6-aws, benchmark-Results/final-thor; tables generated by benchmark-paper/cross_hw/*.py
So far, all efficiency results come from one data-center GPU, but weight-only quantization is usually pitched for smaller, memory-limited devices. We therefore repeat the efficiency study, the time breakdown and a smaller paired task-success study on two such platforms (\cref{tab:devices}): an NVIDIA L4 and an NVIDIA Jetson AGX Thor, an edge system-on-chip built for robots. Both run byte-identical model files.

We repeat the isolated efficiency runs of \cref{sec:efficiency} for all 30 configurations, now with both kernel paths for every real W4 method, and the time breakdown of \cref{sec:anatomy}; task success uses a paired subset of 15 configurations, 120 LIBERO episodes and 252 LIBERO-plus tasks on every GPU (setup in \cref{tab:devices,app:hardware}).

\begin{figure*}[t]
    \centering
    \includegraphics[width=\textwidth]{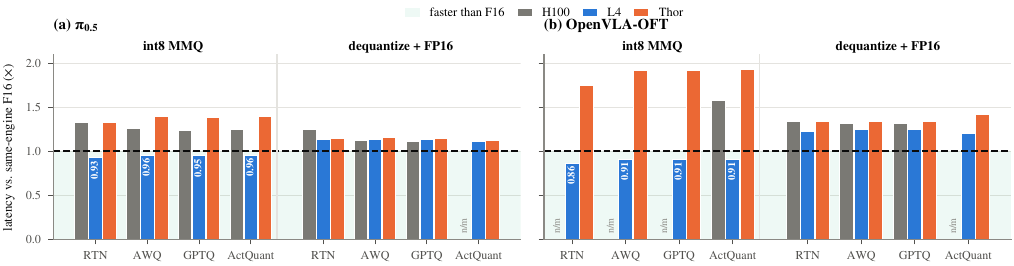}
    \vspace{-5mm}
    \caption{\textbf{The kernel and the GPU decide whether 4 bits are faster.} Latency of every real W4 configuration relative to the same engine's F16. Only int8 kernels on the L4 beat F16 (blue values); the same kernels are the slowest on the Thor, and dequantizing to FP16 is slower than F16 on every GPU. n/m: not measured.}
    \label{fig:cross_hw}
    \vspace{-2mm}
\end{figure*}

% \revComment{run below commented experiment once gpu is free to fill up n/m in Fig 7}
% % Also measure on the H100, same isolated protocol (10 warm-up + 50 calls, exclusive idle GPU):
% % - π0.5 ActQuant 4 bpw on the forced-cuBLAS build of ActQuant's engine
% % - OpenVLA-OFT on ActQuant's OFT engine: RTN Q4_0, AWQ W4, GPTQ W4 on the default MMQ build; ActQuant 4 bpw on the forced-cuBLAS build
% % Use the same build flags as the L4/Thor runs. Output alongside the Q4_0 re-session results.

We summarize the results in \cref{fig:cross_hw}. On the L4, every real W4 method on the int8 path beats F16, by 4--7\% on \pifive{} and 9--14\% on OpenVLA-OFT, the only speed-ups in our study (\cref{fig:cross_hw,tab:cross_e1}); the dequantize path stays slower ($1.11$--$1.25\times$). A plausible reason is that the L4's low bandwidth raises its ridge ($\approx$403\,FLOP/B) above \pifive{}'s prefix ($\approx$350\,FLOP/B), so fewer weight bytes finally help, and its int8 kernels are efficient enough to help even OpenVLA-OFT's compute-bound backbone. Confirming this would need per-operator profiling on the L4, which we leave to future work.
On the edge Thor, every W4 configuration is slower than F16---$1.13$--$1.42\times$ on the dequantize path and up to $1.93\times$ with int8 kernels, the reverse of the L4, and energy per call rises with it. Memory still drops as elsewhere (4-bit \pifive{}: 3.3--3.5 instead of 7.2\,GiB), and again the engine matters more: compiled PyTorch runs \pifive{} $1.5\times$ faster than our F16 engine.
On both the L4 and Thor, \pifive{}'s prefix takes a larger share of GPU time, but the 16-bit action expert still takes 41--45\%.  GR00T's action head 54--59\% (\cref{tab:cross_e2}). The action loop remains a large part of every call on every GPU.

\begin{table}[!htbp]
\centering
\caption{\textbf{Task success on identical episodes across GPUs} (LIBERO: 120 episodes; LIBERO-plus: 252 tasks, paired change vs.\ F16 on the same GPU; bold: $p<0.05$). OpenVLA-OFT: real low-bit at ActQuant's scope. $^\ddagger$Compared with GGML F16, so includes the $\approx$3.7-point engine effect.}

\label{tab:cross_e3}
\scriptsize
\setlength{\tabcolsep}{1.6pt}
\resizebox{\columnwidth}{!}{%
\begin{tabular}{@{}l r r r r r r@{}}
\toprule
 & \multicolumn{3}{c}{LIBERO (\%)} & \multicolumn{3}{c}{LIBERO-plus (\%)}\\
\cmidrule(lr){2-4}\cmidrule(lr){5-7}
Configuration & H100 & L4 & Thor & H100 & L4 & Thor\\
\midrule
\grouprow{7}{\pifive{} (real)}
F16 & 95.0 & 94.2 & 97.5 & 87.3 & 86.5 & 87.3\\
RTN Q4\_0 & 95.0 & 99.2 & 96.7 & 88.5 (+1.2) & 86.9 (+0.4) & 85.7 ($-$1.6)\\
AWQ W4 & 96.7 & 95.8 & 97.5 & 87.3 (+0.0) & 85.3 ($-$1.2) & 86.5 ($-$0.8)\\
GPTQ W4 & 97.5 & 97.5 & 95.8 & 87.3 (+0.0) & 84.5 ($-$2.0) & 86.5 ($-$0.8)\\
ActQuant 4\,bpw$^\ddagger$ & 98.3 & 99.2 & 97.5 & 81.7 (\textbf{$-$5.6}) & 80.6 (\textbf{$-$6.0}) & 79.0 (\textbf{$-$8.3})\\
\grouprow{7}{OpenVLA-OFT (real)}
F16 & 97.5 & 96.7 & 97.5 & 72.2 & 71.8 & 73.0\\
RTN Q4\_0 & 95.0 & 94.2 & 96.7 & 70.6 ($-$1.6) & 69.4 ($-$2.4) & 69.8 ($-$3.2)\\
AWQ W4 & 94.2 & 94.2 & 95.0 & 71.8 ($-$0.4) & 71.8 (+0.0) & 71.8 ($-$1.2)\\
GPTQ W4 & 96.7 & 95.0 & 95.0 & 74.6 (+2.4) & 73.0 (+1.2) & 72.6 ($-$0.4)\\
ActQuant 4\,bpw & 92.5 & 94.2 & 94.2 & 67.1 (\textbf{$-$5.2}) & 68.3 ($-$3.6) & 69.4 ($-$3.6)\\
\grouprow{7}{GR00T N1.5 (sim.)}
FP & 92.5 & 90.8 & 91.7 & 77.0 & 77.8 & 78.6\\
RTN W4 & 90.8 & 90.0 & 95.0 & 79.0 (+2.0) & 77.0 ($-$0.8) & 79.4 (+0.8)\\
QuantVLA W4A8 & 91.7 & 90.8 & 90.0 & 75.0 ($-$2.0) & 76.6 ($-$1.2) & 73.8 ($-$4.8)\\
\grouprow{7}{UniVLA (sim.)}
FP & 90.0 & 92.5 & 91.7 & 54.8 & 54.4 & 55.2\\
RTN W4 & 92.5 & 90.0 & 91.7 & 52.0 ($-$2.8) & 54.8 (+0.4) & 52.0 ($-$3.2)\\
\bottomrule
\end{tabular}}
\vspace{-3mm}
\end{table}

\textbf{Task success does not depend on the GPU.} On identical episodes, the three GPUs agree on 90--99\% of outcomes (\cref{tab:cross_e3}). LIBERO stays saturated everywhere, and weight-only W4 methods stay within 3.2 points of FP, none significantly. ActQuant on OpenVLA-OFT, against its own engine's F16, again loses under perturbation on every GPU (3.6--5.2 points, significant on the H100), matching the full H100 run ($-4.9$; \cref{tab:robust}). For \pifive{}, the subset has no same-engine baseline. With only 252 tasks, the smaller effects of \cref{sec:robust} show up but are not significant with QuantVLA losing on all three GPUs ($-1.2$ to $-4.8$) and 4-bit UniVLA on two.

\takeaway{The same 4-bit files cut memory by $2$--$3\times$ on all three GPUs and leave task success unchanged, but their speed depends on the hardware: up to 14\% faster with int8 kernels on the L4, slower on the H100, and up to $1.9\times$ slower on the edge Thor, where those same int8 kernels are the slowest path. Speed claims must name the GPU and the kernel path.}

\section{Discussion}
\label{sec:discussion}
    \paragraph{Implications for method design.}
Where quantization can pay off depends on the model.  In \pifive{} the action expert is the only memory-bound component (63\% of GPU time), yet most available PTQ methods leave it unquantized. Weights alone does not suffice as the action expert's weight multiplications take only 10.5 of its 37.5\,ms (at most $1.21\times$ faster GPU time), so its other 27\,ms must be fused too. In addition, compute-bound backbones (\pifive{}'s prefix, OpenVLA-OFT's LLM) speed up only if weights and activations both run in INT8 or FP8 on cuBLAS-class kernels. Only a backbone that decodes token by token, as in UniVLA, suits weight-only quantization. Finally, host and launch overhead takes 71 of 131\,ms per eager call, and compiling the unquantized policy alone makes it $3.8\times$ faster, more than any quantizer.
Weight-only quantization is faster only when three things line up: the layers are memory-bound on that GPU, the low-bit kernel runs fast there, and the batch is small. Of our three GPUs, only the L4 comes close, and only with int8 kernels (\cref{sec:hardware}).

\paragraph{Evaluate on a large number of episodes.} With QuantVLA's 200 episodes, only differences above 9.4 points can be detected reliably, so its reported 1.5-point gain could be noise. Our 2{,}000 LIBERO episodes detect differences of 1.6 points, and our paired LIBERO-plus tests 1.9--2.5 points. So when we report no loss, any real loss is most likely smaller than 2--2.5 points.

% \paragraph{Six things we recommend VLA quantization papers report.}
% Each addresses a gap we hit: (i) \emph{the kernel}, since the same AWQ weights ran $1.12$ to $6.8\times$ slower depending on how they executed; (ii) \emph{real or simulated execution}, since simulation cannot show a speed-up; (iii) \emph{same-engine timing}, with GPU and batch size, since the engine alone changed latency by $2.1\times$; (iv) \emph{enough episodes}, about 500 per suite, or fewer when paired with thousands of perturbed tasks (our UniVLA and OpenVLA-OFT runs use 250, following QVLA); (v) \emph{paired tests under distribution shift}, since LIBERO missed losses of up to 4.3 points; and (vi) \emph{generic baselines at the same scope}, since plain AWQ, run under our setup, matched QVLA's reported 96.7\% on OpenVLA-OFT, well above the 92.5\% reported for AWQ.
% Each addresses a gap we hit: (i) \emph{the kernel}, since the same AWQ weights ran $1.12$ to $6.8\times$ slower depending on it; (ii) \emph{real or simulated execution}, since simulation cannot show a speed-up; (iii) \emph{same-engine timing}, with GPU and batch size, since the engine alone changed latency by $2.1\times$; (iv) \emph{enough episodes}, about 500 per suite or thousands of paired perturbed tasks; (v) \emph{paired tests under distribution shift}, since LIBERO missed losses of up to 4.3 points; and (vi) \emph{generic baselines at the same scope}, since plain AWQ matched QVLA's reported 96.7\%, well above the 92.5\% it reports for AWQ.

\subsection{Threats to validity}
\label{sec:audit}
    \begin{table}[!htbp]
\centering
\caption{\textbf{What each VLA-quantization release lets a reader verify} (checked October 2026). Real low-bit: low-bit weights and kernels whose memory and speed can be measured.}
\label{tab:audit}
\scriptsize
\setlength{\tabcolsep}{2pt}
\begin{tabularx}{\columnwidth}{@{}l c X@{}}
\toprule
Method & Real low-bit & What we found\\
\midrule
ActQuant & \cmark & Reproduces its LIBERO results (96.6/96.0/94.5 vs.\ 96.6/96.3/94.8\%), but runs $1.25$--$1.37\times$ slower than F16 on its own engine; its reported speed-up (up to $1.5\times$) is measured on OpenVLA against PyTorch. Model cards name a different base checkpoint than the files were built from.\\
QuantVLA & \xmark & GR00T code only. Simulated W4A8 with floating-point weight copies, so larger and slower; memory computed from bit widths, $1.22\times$ speed-up only in the repository; activation scales frozen after the first request.\\
QVLA & \xmark & Simulated. Code needs fixes to run on OpenVLA-OFT; the released sensitivity proxy is not the paper's estimator and does not regenerate its results.\\
HoloQ-VLA & \xmark & Inspected, not run. Published checkpoints are FP16 with no INT4 kernel (simulated W4A4); the \pifive{} release quantizes 252 of the paper's 414 layers.\\
\bottomrule
\end{tabularx}
\vspace{-3mm}
\end{table}
Our speed results hold for batch 1 on three GPUs; larger batches or fast INT8/FP8 weight-activation kernels could change them. GR00T and UniVLA use simulated quantization, so they speak to task success only, and at 3 bits we compare ActQuant only with RTN on its own int8-activation path. We cannot pin ActQuant's or QuantVLA's robustness loss on activation quantization alone, and part of the engine gap remains unexplained. The cross-GPU subset (252 LIBERO-plus tasks) confirms only losses of about 5 points or more.
\cref{tab:audit} summarizes what each VLA-specific PQT release we consider supports. Three things stand out.

\begin{enumerate}
    \item \textbf{Only one release can be timed.} ActQuant is the only VLA PQT method that ships real low-bit files and reproduces its LIBERO results, but those files run slower than its own 16-bit model. QuantVLA and QVLA only simulate quantization in their published checkpoints. These are stored in 16 bits without a low-bit kernel, so none of them can be timed.

    \item \textbf{What is released is not always what was evaluated.} QuantVLA reports results on \pifive{} but releases code only for GR00T, and QVLA's released code uses a different sensitivity measure from the one described in its paper, which does not reproduce its reported results.
    
    \item \textbf{Small samples and weak baselines can create gains that are not there.} QuantVLA's 1.5-point gain over FP on GR00T comes from 200 episodes; on 2{,}000 we measure $-0.2$, and a 4.2-point loss under distribution shift (\cref{sec:robust}). At QVLA's own scope, plain AWQ reaches 96.6\% on OpenVLA-OFT, level with QVLA's 96.7\% and well above the 92.5\% QVLA reports for AWQ (\cref{app:oft}).
\end{enumerate}

% \takeaway{Of the releases we ran, only ActQuant can be timed, and it is slower than F16 on its own engine; QuantVLA and QVLA only simulate quantization, so their speed cannot be measured. Their reported gains vanish with large paired samples and same-scope baselines. A quantization claim is only as strong as what its release lets others measure.}

\section{Conclusion}
\label{sec:conclusion}
    % Across four VLAs, three GPUs and more than 138{,}000 closed-loop episodes, \bench{} shows that quantization reliably saves memory but rarely saves time, and that standard LIBERO hides the losses it causes. \textcolor{red}{At 4 bits and above}, we found no advantage of VLA-specific post-training quantization over generic methods.

% RTN, AWQ and GPTQ matched the specialized methods on LIBERO and were more robust under distribution shift, where ActQuant and QuantVLA lost 2.3--5.3 points that LIBERO did not reveal; calibration did not improve on plain RTN in our tests, and only UniVLA, with discrete action tokens, lost at 4 bits. Memory fell by up to $2.3\times$, but latency did not, except with int8 kernels on the L4 (4--14\%): in \pifive{}, the quantized backbone is compute-bound while the memory-bound action expert stays in 16 bits.

% Because where the time goes differs across models, future methods should start from profiling, target the components that dominate each model's latency, and report same-engine latency, paired robustness tests and same-scope generic baselines. We will release \bench{} to support such evaluations.

Across four VLAs, three GPUs and more than 180{,}000 closed-loop episodes, \bench{} shows that quantization saves memory but rarely improves latency, while standard LIBERO hides important robustness losses. At 3 and 4 bits, we find no clear advantage of VLA-specific PTQ over generic methods. At 4 bits, RTN, AWQ and GPTQ match the specialized methods on LIBERO and are more robust under distribution shift, where ActQuant and QuantVLA lose 2.3--4.2 points that LIBERO misses. At 3 bits, plain RTN and ActQuant both lose about 5 points, which LIBERO underestimates. Calibration offers no measurable benefit over plain RTN, and UniVLA is the only model where generic 4-bit weights cause significant robustness losses. Real low-bit quantization reduces peak GPU memory by $2$--$3\times$ but does not improve latency on the H100 or the Jetson AGX Thor; only int8 kernels on the L4 achieve 4--14\% speed-ups. On \pifive{}, the quantized backbone is compute-bound, while the memory-bound action expert remains in 16 bits. These findings suggest that effective VLA quantization starts from profiling each model's actual bottlenecks rather than simply reducing weight precision. Future evaluations should report same-engine latency, paired robustness tests and generic baselines at identical scope. We will release \bench{} to support such comparisons.

% \paragraph{Future work.} Our speed results use llama.cpp kernels at batch 1. Next, we will test FP8 and INT8 kernels in TensorRT-LLM and Marlin W4A16 kernels \citep{frantar2025marlin}, quantize and fuse the
% action expert, remove host overhead with CUDA graphs, and run real low-bit GR00T and UniVLA at larger batch sizes.

\bibliography{main}
\bibliographystyle{plainnat}

\appendix
% =============================================================================
    \onecolumn

\section{Quantization Schemes}
\label{app:schemes}
\Cref{tab:taxonomy} lists every quantization scheme in \bench{}: weight and activation precision, calibration data, storage format, and whether execution is real or simulated.
\begin{table}[!htbp]
    \centering
    \caption{\textbf{Quantization schemes in \bench{}.} W/A: bits of weights/activations in matrix multiplications; A8-dyn: int8 activation scale computed per call. Calib.: calibration data (fr.: frames; ep.: episodes). Real: low-bit weights run by low-bit kernels; sim.: rounded weights computed in floating point. bpw includes scales.}
    \label{tab:taxonomy}
    \scriptsize
    \setlength{\tabcolsep}{2.2pt}
    \resizebox{0.8\textwidth}{!}{%
    \begin{tabular}{@{}l l l l l@{}}
    \toprule
    Scheme & Methods & Calib. & Format & Exec.\\
    \midrule
    W8A16 & RTN Q8\_0 & none & int8, block 32, 8.5\,bpw$^a$ & real / sim.\\
    W4A16 & RTN Q4\_0 & none & int4, block 32, 4.5\,bpw$^a$ & real / sim.\\
    W4A16 & AWQ, GPTQ & 512 fr. & uint4, group 128, 4.19\,bpw; run as Q4\_1, 5.0\,bpw$^b$ & real / sim.\\
    W$\cdot$A8-dyn & RTN/AWQ/GPTQ on MMQ & none & int8 act.\ per 32 values & real\\
    W3--4 A8-dyn & ActQuant 4/3.5/3 & 60 ep. & k-quant mix, 3--4\,bpw nominal & real\\
    W4A8 static & QuantVLA & first requests & DuQuant W4, A8 per channel & sim.\\
    \bottomrule
    \end{tabular}}\\[1pt]

    {\scriptsize $^a$On UniVLA and OpenVLA-OFT, RTN uses the baseline formats of QVLA's paper: uint8 per output channel (W8), uint4 group 128 (W4). $^b$Q4\_1 is a lossless llama.cpp re-encoding (identical codes) for the cuBLAS and MMQ paths; the 4.19\,bpw file runs only on our fused kernel.}
    \vspace{-3mm}

\end{table}

\section{Roofline of the Unquantized and Quantized Policies}
\label{app:roofline}

\begin{figure}[!htbp]
    \centering
    \includegraphics[width=\columnwidth]{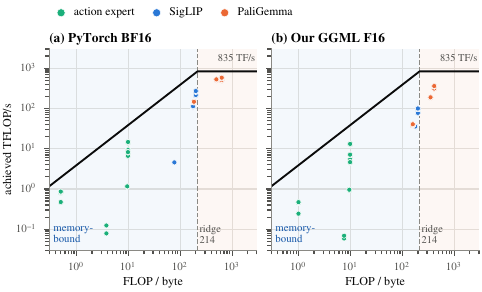}
    \vspace{-2mm}
    % \caption{\textbf{Every weight multiplication of the unquantized \pifive{} on the H100 roofline}: (a) PyTorch BF16, (b) our GGML F16 engine. The action expert (green) is memory-bound, PaliGemma (orange) compute-bound and SigLIP (blue) near the ridge. Intensity is analytic; TFLOP/s is analytic FLOPs over measured kernel time.}
    \caption{\textbf{Only the untargeted action expert is memory-bound} (unquantized \pifive{}, H100 roofline). Each dot is one weight multiplication: the action expert (green) is memory-bound, PaliGemma (orange) compute-bound and SigLIP (blue) just below the ridge, in both (a) PyTorch and (b) our GGML engine.}
    \vspace{-3mm}
    \label{fig:roofline}
\end{figure}

\begin{figure}[!htbp]
\centering
\includegraphics[width=0.9\textwidth]{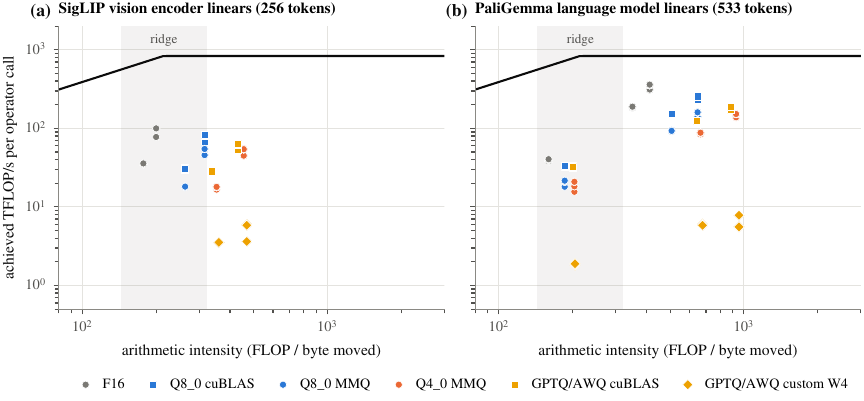}
\caption{GGML SigLIP and PaliGemma GEMMs before and after quantization. Points move right (intensity computed on packed weight bytes) and down (lower achieved TFLOP/s). For the dequantize path the rightward shift is nominal: the GEMM itself still reads FP16 weights after an extra dequantization pass.}
\label{fig:qroofline}
\end{figure}

\begin{table}[!htbp]
\centering
\caption{\textbf{Not one quantized layer is faster than F16.} GPU time of exactly the quantized layers relative to the same layers in F16 on the same engine, and the time added per call. Faster: quantized operators faster than F16. All methods quantize the same 288 linears (plus the embedding); our GGML engine skips 5 of them in the last PaliGemma layer, whose output the policy never uses, hence 283.}

\label{tab:slowdown}
\scriptsize
\setlength{\tabcolsep}{2.4pt}
\resizebox{0.7\textwidth}{!}{%
\begin{tabular}{@{}l c c c r c@{}}
\toprule
Configuration & SigLIP & PaliGemma & Total & +ms & Faster\\
\midrule
Q8\_0 cuBLAS & 1.16$\times$ & 1.33$\times$ & 1.24$\times$ & 3.4 & 0/283\\
Q8\_0 MMQ & 1.85$\times$ & 2.17$\times$ & 2.00$\times$ & 14.1 & 0/283\\
Q4\_0 cuBLAS / MMQ & 1.36 / 1.88$\times$ & 1.79 / 2.26$\times$ & 1.57 / 2.06$\times$ & 8.0 / 14.9 & 0/283\\
AWQ, GPTQ cuBLAS & 1.36$\times$ & 1.78$\times$ & 1.56$\times$ & 7.8 & 0/283\\
AWQ, GPTQ MMQ & 1.96$\times$ & 2.49$\times$ & 2.21$\times$ & 17.1 & 0/283\\
AWQ, GPTQ fused & 14.0$\times$ & 44.8$\times$ & 28.7$\times$ & 389.5 & 0/283\\
ActQuant 4 bpw & 1.96$\times$ & 2.03$\times$ & 2.01$\times$ & 20.8 & 0/288\\
ActQuant 3.5 bpw & 2.19$\times$ & 2.36$\times$ & 2.31$\times$ & 27.1 & 0/288\\
ActQuant 3 bpw & 2.19$\times$ & 2.57$\times$ & 2.44$\times$ & 29.9 & 0/288\\
\bottomrule
\end{tabular}}
\vspace{-3mm}
\end{table}

\begin{table}[!htbp]
\centering
\caption{Where the official \pifive{} BF16 policy spends GPU time (OpenPI PyTorch, eager, batch 1, 10 flow steps). AI: median arithmetic intensity of the component's weight GEMMs. TFLOP/s, GB/s and \% of roof: median achieved values (time-weighted TFLOP/s in parentheses); peaks are datasheet values. PyTorch runs SigLIP on three image slots (two cameras, one masked) and pads the prompt to 200 tokens, giving a 968-token prefix. The total also includes the projector and kernels outside the three components (1.0\,ms). GEMM includes SigLIP's patch-embedding convolution; Figure~\ref{fig:anatomy} shows only the quantizable linears (11\,ms).
}
\scriptsize
\begin{tabular}{@{}l c c c r r r r l r r r@{}}
\toprule
Component & Quantized & Tokens & Calls & GPU ms (share) & GEMM & Non-GEMM & AI & Class & TFLOP/s & GB/s & \% roof\\
\midrule
SigLIP & yes & 256 & 3 & 7.6 (13\%) & 4.0 & 3.6 & 177 & near ridge & 110 (168) & 621 & 16\%\\
PaliGemma & yes & 968 & 1 & 13.2 (22\%) & 7.2 & 5.9 & 498 & compute-bound & 509 (530) & 904 & 61\%\\
Action expert & no & 10 & 10 & 37.5 (63\%) & 10.5 & 27.0 & 10 & memory-bound & 7 (6) & 931 & 24\%\\
\midrule
Total & & & & 59.4 & 21.8 & 37.6 & & & & & \\
\bottomrule
\end{tabular}
\end{table}

On our GGML F16 engine the classes are the same: action-expert GEMMs are memory-bound (median 469\,GB/s, 12\% of roof, 15.2\,ms), PaliGemma (533 tokens) is compute-bound at 22\% of roof (median; 36\% time-weighted), and SigLIP sits near the ridge at 5\%; GPU time is 53.9\,ms (SigLIP 11.6, PaliGemma 12.0, action expert 29.8) plus 18.3\,ms host time.

\section{Per-Suite LIBERO Success of \pifive{}}
\label{app:libero}

% \Cref{fig:libero} shows the success of every \pifive{} configuration with 95\% confidence intervals, per suite. Nearly all intervals overlap the same-engine baseline; the two exceptions are AWQ on MMQ (Goal, FP16 overflow) and ActQuant 3\,bpw (Long).

% \begin{figure}[!htbp]
% \centering
% \includegraphics[width=0.95\textwidth]{figures/f03_pi05_libero_success.pdf}
% \caption{\textbf{\pifive{} LIBERO success} with Wilson 95\% CIs, overall and per suite; dashed lines mark the same-engine unquantized baseline. Colour encodes the method, marker the kernel path (circle: int8 MMQ; square: dequantize + FP16 GEMM; diamond: fused W4A16; triangle: ActQuant engine; hollow: reference only).}
% \label{fig:libero}
% \end{figure}
\Cref{tab:pi05} reports all 18 \pifive{} configurations on LIBERO, with per-suite success, latency and memory. Nearly all match their same-engine baseline; the exceptions are AWQ on MMQ (FP16 overflow) and the 3-bit models on the int8-activation path.

\begin{table}[!htbp]
    \centering
    \caption{\textbf{\pifive{} on LIBERO: quantized models match their baseline (except AWQ on MMQ and the 3-bit int8-activation models), and none is faster than its own engine's 16-bit model} (2{,}000 episodes each). Latency is the closed-loop median per call; $\times$base is relative to that engine's unquantized model (grey row). Peak memory is device-wide, including the simulator. $^{\ddagger}$One FP16 overflow in AWQ closed the server; the 68 affected episodes count as failures (97.7\% on the 432 executed). $^{\S}$ActQuant's own F16 release, built from a different (LeRobot) checkpoint than its quantized files; reference only. Row 18 reports task success only (n/m: closed-loop latency and memory not measured); its isolated speed is in \cref{tab:isolated}.}
    \label{tab:pi05}
    \scriptsize
    \setlength{\tabcolsep}{3.2pt}
    \resizebox{\textwidth}{!}{%
    \begin{tabular}{@{}r l l c c c c c c r c c@{}}
        \toprule
        \# & Method & Scheme (path) & File GB & Success \% (95\% CI) & Spatial & Object & Goal & Long & Lat.\ ms & $\times$base & Peak GiB\\
        \midrule
        \grouprow{12}{GGML engine (ours)}
        \rowcolor{grp!50} 1 & F16 & W16A16 & 6.71 & 96.5 (95.6--97.2) & 98.4 & 99.2 & 97.0 & 91.4 & 81 & 1.00 & 8.36\\
        2 & RTN Q8\_0 & W8A8-dyn (MMQ) & 3.97 & 96.7 (95.8--97.4) & 98.2 & 98.4 & 97.0 & 93.0 & 102 & 1.27 & 5.81\\
        3 & RTN Q8\_0 & W8A16 (cuBLAS) & 3.97 & 96.7 (95.8--97.4) & 98.8 & 97.8 & 98.0 & 92.0 & 94 & 1.17 & 5.87\\
        16 & RTN Q4\_0 & W4A8-dyn (MMQ) & 2.51 & 96.8 (95.9--97.5) & 97.8 & 98.0 & 97.4 & 94.0 & 113 & 1.39 & 4.44\\
        17 & RTN Q4\_0 & W4A16 (cuBLAS) & 2.51 & 96.9 (96.0--97.6) & 99.0 & 98.2 & 97.8 & 92.6 & 93 & 1.15 & 4.51\\
        4 & AWQ & W4A8-dyn (MMQ) & 2.70 & 93.5 (92.3--94.5) & 98.2 & 98.4 & 84.4$^{\ddagger}$ & 92.8 & 113 & 1.40 & 4.62\\
        5 & AWQ & W4A16 (cuBLAS) & 2.70 & 97.0 (96.1--97.6) & 98.6 & 99.4 & 97.6 & 92.2 & 95 & 1.17 & 4.68\\
        6 & AWQ & W4A16 (fused) & 2.40 & 97.2 (96.3--97.8) & 99.4 & 98.2 & 96.6 & 94.4 & 435 & 5.38 & 4.33\\
        7 & GPTQ & W4A8-dyn (MMQ) & 2.70 & 97.6 (96.8--98.2) & 98.6 & 99.2 & 98.2 & 94.4 & 105 & 1.30 & 4.62\\
        8 & GPTQ & W4A16 (cuBLAS) & 2.70 & 97.1 (96.3--97.7) & 98.6 & 98.8 & 97.0 & 94.0 & 89 & 1.10 & 4.68\\
        9 & GPTQ & W4A16 (fused) & 2.40 & 97.0 (96.2--97.7) & 98.2 & 98.4 & 98.0 & 93.6 & 438 & 5.42 & 4.32\\
        \midrule
        \grouprow{12}{ActQuant engine (released, unmodified)}
        \rowcolor{grp!50} 10 & F16 (our GGUF) & W16A16 & 6.71 & 97.0 (96.1--97.6) & 98.2 & 98.8 & 97.0 & 93.8 & 99 & 1.00 & 8.05\\
        11 & ActQuant & mixed 4\,bpw, A8-dyn & 2.63 & 96.6 (95.7--97.3) & 97.8 & 99.2 & 96.8 & 92.6 & 115 & 1.16 & 4.77\\
        12 & ActQuant & mixed 3.5\,bpw, A8-dyn & 2.47 & 96.0 (95.1--96.8) & 98.6 & 99.0 & 94.6 & 92.0 & 121 & 1.22 & 4.63\\
        13 & ActQuant & mixed 3\,bpw, A8-dyn & 2.36 & 94.5 (93.4--95.4) & 97.0 & 99.0 & 95.0 & 87.0 & 122 & 1.23 & 4.52\\
        18 & RTN Q3\_K & W3A8-dyn (MMQ) & 2.47 & 95.7 (94.7--96.5) & 98.6 & 98.8 & 96.4 & 88.8 & n/m & n/m & n/m\\
        14 & ActQuant own F16$^{\S}$ & W16A16 & 6.71 & 95.2 (94.1--96.0) & 96.6 & 99.0 & 90.2 & 94.8 & 118 & -- & 8.05\\
        \midrule
        \grouprow{12}{Official OpenPI PyTorch}
        \rowcolor{grp!50} 15 & BF16 (\texttt{torch.compile}) & W16A16 & 7.23 & 96.2 (95.2--96.9) & 98.0 & 98.8 & 96.6 & 91.2 & 47 & 1.00 & 9.34\\
        \bottomrule
    \end{tabular}%
    }
    \vspace{-2mm}
\end{table}

\section{Robustness by Difficulty Level}
\label{app:difficulty}

\begin{table}[!htbp]
\centering
\caption{LIBERO-plus success (\%) per difficulty level. UniVLA and OpenVLA-OFT levels from their 50\% subset (tasks without a level label omitted).}
\scriptsize
\resizebox{\textwidth}{!}{%
\begin{tabular}{@{}l l c c c c c@{}}
\toprule
Model & Config & L1 & L2 & L3 & L4 & L5\\
\midrule
\pifive{} & GGML F16 / Q8\_0 / Q4\_0 / AWQ / GPTQ & 97.6 / 97.6 / 95.8 / 95.5 / 96.7 & 94.7 / 94.9 / 94.9 / 95.4 / 95.2 & 90.7 / 93.1 / 91.8 / 92.0 / 90.7 & 83.8 / 84.6 / 85.9 / 85.6 / 85.1 & 65.5 / 67.1 / 64.8 / 62.7 / 63.0\\
\pifive{} & AQ F16 / AQ 4 / AQ 3 bpw & 94.2 / 94.8 / 93.9 & 92.4 / 91.2 / 89.2 & 90.0 / 86.5 / 81.6 & 79.9 / 75.5 / 72.6 & 57.5 / 54.4 / 50.8\\
GR00T & FP / QuantVLA & 87.0 / 88.8 & 87.3 / 87.1 & 83.0 / 79.0 & 75.2 / 70.2 & 56.2 / 43.8\\
UniVLA & FP / W8 / RTN W4 / AWQ / GPTQ & 79.6 / 80.7 / 74.1 / 72.5 / 79.0 & 60.2 / 59.7 / 57.2 / 54.5 / 59.0 & 54.2 / 54.4 / 52.6 / 53.0 / 52.2 & 44.7 / 45.5 / 45.0 / 41.0 / 41.6 & 25.6 / 25.7 / 25.2 / 21.0 / 25.1\\
OpenVLA-OFT & FP / W8 / RTN W4 / AWQ / GPTQ & 92.1 / 91.6 / 90.3 / 90.5 / 92.1 & 90.2 / 90.3 / 88.7 / 89.0 / 89.1 & 77.6 / 78.3 / 77.9 / 77.0 / 78.2 & 69.4 / 69.6 / 67.3 / 69.1 / 69.6 & 37.1 / 38.1 / 37.6 / 36.0 / 36.7\\
\bottomrule
\end{tabular}}
\end{table}

\paragraph{Per-suite effects.} Losses are concentrated rather than spread. On LIBERO-plus, QuantVLA's entire GR00T deficit comes from the Object ($-4.5$) and Goal ($-12.3$) suites, with Spatial and Long unchanged; on UniVLA, RTN W4 loses most on the long-horizon suite ($-6.0$) and AWQ on Object ($-9.3$); OpenVLA-OFT's largest loss is RTN W4 on Object ($-4.0$). On LIBERO itself, ActQuant 3\,bpw already loses 6.8 points on the long-horizon suite, where errors accumulate over up to 520 steps. Suite-level reporting is therefore necessary to diagnose a method: averaged scores can hide a suite-specific collapse.

\section{Flow-Matching Noise Diagnostic}
\label{app:noise}

% This diagnostic splits the 3.7-point LIBERO-plus gap between our GGML engine and ActQuant's engine (same F16 file) into a flow-matching-noise part ($-1.8$) and an engine part ($-1.9$) (\cref{sec:robust}).
Running the same F16 file on ActQuant's engine costs 3.7 points on LIBERO-plus (\cref{sec:robust}). \Cref{tab:noise} gives two exploratory diagnostics, neither significant after Holm correction: fixing the flow-matching noise on our engine, and comparing ActQuant's engine with that fixed-noise run.

\begin{table}[!htbp]
\centering
\caption{Effect of fixed flow-matching noise on \pifive{} F16 (LIBERO-plus, 2{,}016 tasks). \sig paired McNemar $p<0.05$.}
\label{tab:noise}
\scriptsize
\begin{tabular}{@{}l c c c c c c c c@{}}
\toprule
 & All & Camera & Robot & Language & Light & Backgr. & Noise & Layout\\
\midrule
GGML F16, noise varies per call & 86.6 & 75.3 & 77.8 & 86.8 & 95.8 & 92.4 & 90.6 & 87.5\\
GGML F16, fixed noise & 84.8 & 70.8 & 74.0 & 88.9 & 95.1 & 89.6 & 90.6 & 84.7\\
ActQuant-engine F16 (fixed by design) & 82.9 & 69.4 & 71.9 & 82.6 & 94.1 & 91.0 & 86.1 & 85.4\\
\midrule
fixed $-$ varying (paired) & $-1.8$\sig & $-4.5$ & $-3.8$ & $+2.1$ & $-0.7$ & $-2.8$ & $+0.0$ & $-2.8$\\
ActQuant engine $-$ fixed (paired) & $-1.9$\sig & $-1.4$ & $-2.1$ & $-6.2$\sig & $-1.0$ & $+1.4$ & $-4.5$\sig & $+0.7$\\
\bottomrule
\end{tabular}
\end{table}
\paragraph{Why the same F16 file behaves differently.} Identical weights do not imply identical computation. \pifive{} generates each action chunk by refining random noise over 10 flow steps: our engine draws fresh noise on every call, whereas ActQuant's engine reuses one fixed sample, so an uncertain policy can repeat the same unsuccessful action instead of varying it. The engines also preprocess images and tokenize instructions in their own code, and their kernels accumulate in different orders. On LIBERO the policy is confident and these small differences rarely change the outcome; under LIBERO-plus perturbations it is often uncertain, and in the closed loop a slightly different action changes the next observation, so small differences compound over hundreds of steps. Fixed noise accounts for about half of the 3.7-point gap; the remainder concentrates on language and sensor-noise shifts, consistent with, but not proof of, a preprocessing difference.

\section{Offline Action Fidelity}
\label{app:fidelity}

\begin{table}[!htbp]
\centering
\caption{\textbf{Offline action error does not predict closed-loop success} (\pifive{}; mean relative L2 of the action chunk against an unquantized reference, same frames and noise per engine). RTN, AWQ and GPTQ differ by about 20$\times$ in error yet reach the same success; ActQuant 4\,bpw has an error similar to AWQ's yet loses 2.3 points on LIBERO-plus. ActQuant runs only on its own engine, which draws its own noise, so it is compared with that engine's F16.}
\scriptsize
\setlength{\tabcolsep}{3pt}
\begin{tabular}{@{}l c c c@{}}
\toprule
Method & vs.\ PyTorch FP (16 frames) & vs.\ GGML F16 (MMQ / cuBLAS / fused) & vs.\ ActQuant-engine F16\\
\midrule
RTN Q8\_0 & 0.58\% & 0.21\% / 0.15\% / -- & --\\
GPTQ W4 & 2.05\% (linears) / 2.96\% (+ emb.) & 2.26\% / 2.21\% / 2.21\% & --\\
AWQ W4 & 3.03\% & 3.63\% / 3.63\% / 3.61\% & --\\
RTN Q4\_0 & 12.0\% (one outlier frame; median 2.0\%) & 2.49\% / 2.46\% / -- & --\\
ActQuant 4 / 3.5 / 3\,bpw & -- & -- & 3.3\% / 8.1\% / 8.8\%\\
\bottomrule
\end{tabular}
\end{table}

GR00T action error (relative L2, 6 fixed frames, all 8 denoising steps; not comparable with \pifive{}): QuantVLA 14.95\% mean / 28.04\% max; RTN Q8\_0 0.33 / 0.74; RTN Q4\_0 1.59 / 3.33; AWQ 1.67 / 2.65; GPTQ 0.49 / 0.66.

\section{Calibrated Methods vs.\ RTN on LIBERO-plus}
\label{app:calib}

\Cref{tab:calib} compares AWQ and GPTQ directly with plain RTN at the same bit width, paired on identical LIBERO-plus tasks. No calibrated method is significantly more robust than RTN on any model.

\begin{table}[!htbp]
\centering
\caption{Paired change in LIBERO-plus success of AWQ and GPTQ relative to RTN W4 (points; exact McNemar $p$). Positive: the calibrated method is more robust than RTN.}
\label{tab:calib}
\scriptsize
\begin{tabular}{@{}l c c c@{}}
\toprule
Model & Tasks & AWQ $-$ RTN & GPTQ $-$ RTN\\
\midrule
\pifive{} & 2{,}016 & $-$0.4 ($p=0.64$) & $-$0.5 ($p=0.56$)\\
GR00T N1.5 & 2{,}016 & +0.3 ($p=0.78$) & $-$0.6 ($p=0.45$)\\
UniVLA & 5{,}014 & $\mathbf{-2.4}$ ($p<10^{-3}$) & +0.5 ($p=0.38$)\\
OpenVLA-OFT & 5{,}014 & $-$0.1 ($p=0.93$) & +0.7 ($p=0.11$)\\
\bottomrule
\end{tabular}
\end{table}

\section{GR00T N1.5 Details}
\label{app:groot}

Per-suite LIBERO and LIBERO-plus success of every GR00T configuration, and the paired change per LIBERO-plus perturbation category (\cref{sec:robust}).

\begin{table}[!htbp]
\centering
\caption{GR00T N1.5 per suite and per LIBERO-plus category. Category columns: success of FP, and paired $\Delta$ for the others (\sig $p<0.05$, uncorrected).}
\scriptsize
\setlength{\tabcolsep}{3pt}
\begin{tabular}{@{}l c c c c c c c c c c c c c c c@{}}
\toprule
 & \multicolumn{4}{c}{LIBERO (Sp/Ob/Go/Lo)} & \multicolumn{4}{c}{LIBERO-plus (Sp/Ob/Go/Lo)} & \multicolumn{7}{c}{LIBERO-plus category}\\
\cmidrule(lr){2-5}\cmidrule(lr){6-9}\cmidrule(lr){10-16}
Method & Sp & Ob & Go & Lo & Sp & Ob & Go & Lo & Cam. & Robot & Lang. & Light & Backg. & Noise & Layout\\
\midrule
FP & 95.6 & 95.8 & 94.8 & 86.0 & 83.9 & 81.5 & 75.8 & 71.6 & 72.2 & 45.5 & 83.7 & 94.8 & 94.4 & 89.2 & 67.7\\
QuantVLA & 96.2 & 97.6 & 92.0 & 85.2 & 83.7 & 77.0 & 63.5 & 71.8 & $-$10.8\sig & $-$5.2\sig & $-$6.6\sig & $-$5.6\sig & $-$4.2 & $-$2.1 & +4.9\\
Q8\_0 & 94.2 & 95.6 & 95.6 & 86.8 & 84.3 & 83.7 & 74.2 & 71.6 & $-$0.3 & $-$1.4 & $-$0.3 & $-$1.0 & $-$0.3 & +3.1 & +2.1\\
Q4\_0 & 95.6 & 96.0 & 94.6 & 86.2 & 81.7 & 81.5 & 75.8 & 73.8 & $-$1.7 & $-$0.7 & $-$1.7 & $-$2.4 & $-$0.7 & +2.8 & +4.5\sig\\
AWQ & 95.4 & 95.6 & 94.0 & 85.8 & 81.3 & 84.1 & 75.6 & 72.8 & 0.0 & +1.0 & +1.4 & $-$1.4 & $-$2.4 & +0.3 & +2.8\\
GPTQ & 97.0 & 95.0 & 94.8 & 86.4 & 82.5 & 82.1 & 74.8 & 71.0 & $-$0.7 & $-$1.7 & +1.7 & $-$4.2\sig & $-$2.1 & +2.1 & +0.7\\
\bottomrule
\end{tabular}
\end{table}

\section{OpenVLA-OFT and UniVLA Details}
\label{app:oft}

\begin{table}[!htbp]
\centering
\caption{LIBERO success (\%) for UniVLA and OpenVLA-OFT with simulated weight-only quantization at QVLA's 435-module scope, 25 trials per task (1{,}000 episodes). Weight storage is calculated from the quantized format, not measured. QVLA paper values (OpenVLA-OFT, Table 2 of \citet{xu2026qvla}) are the authors' numbers under their protocol, shown for context only.}
\scriptsize
\begin{tabular}{@{}l l c c c c c c@{}}
\toprule
Model & Configuration & Spatial & Object & Goal & Long & Avg (95\% CI) & Storage GB\\
\midrule
\multirow{5}{*}{UniVLA} & FP (BF16) & 95.2 & 94.8 & 90.8 & 90.0 & 92.7 (90.9--94.2) & 15.11\\
 & RTN W8 (per channel) & 96.0 & 94.4 & 92.8 & 92.0 & 93.8 (92.1--95.1) & 7.91\\
 & RTN W4 (g128) & 97.6 & 91.6 & 93.2 & 86.0 & 92.1 (90.3--93.6) & 4.47\\
 & AWQ W4 (g128) & 94.8 & 94.0 & 91.2 & 88.0 & 92.0 (90.2--93.5) & 4.47\\
 & GPTQ W4 (g128) & 96.0 & 96.0 & 94.0 & 90.8 & 94.2 (92.6--95.5) & 4.47\\
\midrule
\multirow{5}{*}{OpenVLA-OFT (ours)} & FP (BF16) & 98.8 & 96.8 & 99.6 & 94.8 & 97.5 (96.3--98.3) & 15.42\\
 & RTN W8 (per channel) & 98.4 & 98.0 & 98.4 & 96.0 & 97.7 (96.6--98.5) & 8.22\\
 & RTN W4 (g128) & 96.8 & 97.2 & 98.4 & 92.8 & 96.3 (94.9--97.3) & 4.78\\
 & AWQ W4 (g128) & 96.0 & 98.4 & 98.8 & 93.2 & 96.6 (95.3--97.6) & 4.78\\
 & GPTQ W4 (g128) & 97.6 & 98.4 & 98.0 & 91.2 & 96.3 (94.9--97.3) & 4.78\\
\midrule
\multirow{3}{*}{OpenVLA-OFT (QVLA paper)} & FP & 97.6 & 98.4 & 97.9 & 94.5 & 97.1 & 15.4\\
 & W8A16: QVLA / AWQ & & & & & 97.0 / 95.1 & 7.4 / 8.0\\
 & W4A16: QVLA / AWQ & & & & & 96.7 / 92.5 & 4.5 / 5.2\\
\bottomrule
\end{tabular}
\end{table}

UniVLA LIBERO-plus (5{,}014 paired tasks): the full-precision policy is extremely brittle to camera viewpoint (5.0\%) and sensor noise (17.7\%), so floor effects limit what those categories can reveal. OpenVLA-OFT on the same subset reaches 73.1\% (FP), 73.4\% (RTN W8), 72.2\% (RTN W4), 72.1\% (AWQ W4) and 72.9\% (GPTQ W4). Its hardest categories are robot initial state (33.6--35.5\% for every configuration) and camera viewpoint (62.0--64.1\%), and quantization leaves both unchanged; the only large category effect is lighting under RTN W4 (89.5\% vs.\ 94.6\%). The OpenVLA-OFT full-precision run used 50 trials per task; for paired and unpaired comparisons we use its first 25 trials so that all configurations share the same episodes.

\section{Cross-Hardware Details (L4 and Jetson AGX Thor)}
\label{app:hardware}

\paragraph{Setup differences.} Both platforms run the H100's model files byte for byte (checked by sha256) and the same configurations, scripts and seeds. \emph{L4} (AWS g6.2xlarge): Ubuntu 24.04, driver 580, the H100 server's Python environments copied unchanged, all GGML/ActQuant engines rebuilt for sm\_89 with identical CMake flags; the GPU runs at its 72\,W cap. \emph{Jetson AGX Thor}: L4T R39.2.1 (JetPack 7), CUDA 13.2, power mode MAXN with \texttt{jetson\_clocks} (GPU 1575\,MHz, memory 4266\,MHz, fan at maximum; no throttling, junction $\leq$56\,$^\circ$C), desktop stopped during measurements. Its aarch64/CUDA-13 platform forces newer software: PyTorch 2.9.1 for every model (H100: 2.2--2.7), flash-attention 2.8.3 built for sm\_110 (GR00T, UniVLA), TensorFlow 2.15 CPU for OpenVLA-OFT image preprocessing, JAX CPU-only in the \pifive{} environment; model-specific libraries (e.g.\ transformers) keep the H100 versions. The UniVLA W4 artifacts regenerated on the Thor are byte-identical to the H100's. \emph{Measurement}: NVML on the Thor has neither an energy counter nor per-process memory, so power is read from the board monitors every 10\,ms (VDD\_GPU, the GPU rail, comparable to the GPU-board energy of the other GPUs; VIN, the whole module) and integrated; memory is the growth of system memory in use, which on unified memory also counts host-side copies and is not comparable with discrete-GPU memory (e.g.\ 13.6\,GiB for the ActQuant engine's F16). In E3, 3--6 jobs shared a GPU, so only success is used.

\begin{table}[!htbp]
    \centering
    \caption{\textbf{Three platforms, from data center to robot.} Ridge: dense FP16 throughput / memory bandwidth (FLOP/B, datasheets); above it, a layer is compute-bound. Energy: GPU-board counter on H100 and L4; GPU rail and whole module on the Thor. $^a$No published dense FP16 figure.}
    \label{tab:devices}
    \scriptsize
    \setlength{\tabcolsep}{2pt}
    \resizebox{0.7\textwidth}{!}{%
        \begin{tabular}{@{}l l l l@{}}
        \toprule
         & H100 NVL & L4 (AWS g6.2xlarge) & Jetson AGX Thor\\
        \midrule
        Architecture & Hopper (sm\_90) & Ada (sm\_89) & Blackwell (sm\_110), integrated GPU\\
        Memory & 94\,GB, 3.9\,TB/s & 24\,GB, 0.30\,TB/s & 128\,GB unified, 0.27\,TB/s\\
        FP16 TFLOP/s / ridge & 835 / 214 & 121 / 403 & --$^a$\\
        Power & 400\,W limit & 72\,W limit & MAXN mode, locked clocks\\
        Host CPU & 64 cores & 8 vCPU & 14 Arm cores\\
        CUDA / PyTorch & 12 / 2.2--2.7 & 12 / 2.2--2.7 & 13.2 / 2.9\\
        \bottomrule
        \end{tabular}}
    \vspace{-3mm}
\end{table}

\begin{table}[!htbp]
\centering
\caption{\textbf{Latency per call on three GPUs} (milliseconds per policy call, measured alone on an idle GPU, one request at a time). In parentheses: ratio to the same engine's F16 on the same GPU; bold: faster than F16, which happens only with int8 kernels on the L4. n/m: not measured. Energy and memory in \cref{app:hardware}.}

\label{tab:cross_e1}
\scriptsize
\setlength{\tabcolsep}{2.2pt}
\resizebox{0.7\textwidth}{!}{%
\begin{tabular}{@{}l r r r@{}}
\toprule
Configuration & H100 NVL & L4 & Jetson AGX Thor\\
\midrule
\grouprow{4}{\pifive{} (real low-bit)}
GGML F16 & 67 & 167 & 186\\
RTN Q4\_0 MMQ & 89 (1.33$\times$) & 154 (\textbf{0.93$\times$}) & 247 (1.33$\times$)\\
RTN Q4\_0 cuBLAS & 83 (1.25$\times$) & 189 (1.13$\times$) & 213 (1.15$\times$)\\
AWQ W4 MMQ & 84 (1.26$\times$) & 160 (\textbf{0.96$\times$}) & 260 (1.40$\times$)\\
AWQ W4 cuBLAS & 75 (1.12$\times$) & 190 (1.14$\times$) & 216 (1.16$\times$)\\
GPTQ W4 MMQ & 83 (1.24$\times$) & 159 (\textbf{0.95$\times$}) & 258 (1.38$\times$)\\
GPTQ W4 cuBLAS & 74 (1.11$\times$) & 190 (1.14$\times$) & 213 (1.15$\times$)\\
AQ-engine F16 & 85 & 235 & 281\\
ActQuant 4\,bpw MMQ & 106 (1.25$\times$) & 225 (\textbf{0.96$\times$}) & 392 (1.39$\times$)\\
ActQuant 4\,bpw cuBLAS & n/m & 260 (1.11$\times$) & 318 (1.13$\times$)\\
PyTorch BF16 eager & 124 & 327 & 241\\
PyTorch BF16 compiled & 32 & 158 & 126\\
\grouprow{4}{OpenVLA-OFT (real low-bit)}
AQ-engine F16 & 77 & 286 & 255\\
RTN Q4\_0 MMQ & n/m & 247 (\textbf{0.86$\times$}) & 446 (1.75$\times$)\\
RTN Q4\_0 cuBLAS & 103 (1.34$\times$) & 352 (1.23$\times$) & 341 (1.34$\times$)\\
AWQ W4 MMQ & n/m & 261 (\textbf{0.91$\times$}) & 491 (1.92$\times$)\\
AWQ W4 cuBLAS & 101 (1.32$\times$) & 357 (1.25$\times$) & 342 (1.34$\times$)\\
GPTQ W4 MMQ & n/m & 260 (\textbf{0.91$\times$}) & 490 (1.92$\times$)\\
GPTQ W4 cuBLAS & 102 (1.32$\times$) & 356 (1.25$\times$) & 342 (1.34$\times$)\\
ActQuant 4\,bpw MMQ & 122 (1.58$\times$) & 260 (\textbf{0.91$\times$}) & 492 (1.93$\times$)\\
ActQuant 4\,bpw cuBLAS & n/m & 344 (1.20$\times$) & 361 (1.42$\times$)\\
PyTorch BF16 & 59 & 218 & 182\\
\bottomrule
\end{tabular}}
\vspace{-3mm}
\end{table}

\begin{table}[!htbp]
\centering
\caption{Energy per call (J) and memory (GiB) of the real low-bit configurations. H100 and L4: GPU-board energy (NVML) and per-process peak GPU memory. Thor: GPU-rail (VDD\_GPU) and whole-module (VIN) energy, and unified-memory growth (not comparable with the discrete GPUs). n/m: not measured.}
\label{tab:cross_energy}
\scriptsize
\setlength{\tabcolsep}{3pt}
\begin{tabular}{@{}l r r r r r r r@{}}
\toprule
 & \multicolumn{4}{c}{Energy per call (J)} & \multicolumn{3}{c}{Memory (GiB)}\\
\cmidrule(lr){2-5}\cmidrule(lr){6-8}
Configuration & H100 & L4 & Thor GPU & Thor VIN & H100 & L4 & Thor (unified)\\
\midrule
\grouprow{8}{\pifive{}}
GGML F16 & 13.8 & 11.8 & 7.0 & 14.7 & 7.0 & 6.6 & 7.2\\
RTN Q4\_0 MMQ & 17.5 & 11.0 & 8.5 & 17.1 & 3.1 & 2.7 & 3.3\\
RTN Q4\_0 cuBLAS & 16.5 & 13.4 & 8.0 & 16.9 & 3.2 & 2.7 & 3.3\\
AWQ W4 MMQ & 16.9 & 11.4 & 8.8 & 17.7 & 3.3 & 2.8 & 3.4\\
AWQ W4 cuBLAS & 15.9 & 13.5 & 8.1 & 17.1 & 3.4 & 2.9 & 3.5\\
GPTQ W4 MMQ & 16.8 & 11.3 & 8.7 & 17.5 & 3.3 & 2.8 & 3.4\\
GPTQ W4 cuBLAS & 15.5 & 13.5 & 8.0 & 17.0 & 3.4 & 2.9 & 3.5\\
AQ-engine F16 & 20.9 & 16.8 & 11.6 & 22.9 & 6.7 & 5.7 & 13.6\\
ActQuant 4\,bpw MMQ & 24.8 & 16.1 & 14.7 & 28.4 & 3.4 & 2.3 & 5.9\\
ActQuant 4\,bpw cuBLAS & n/m & 18.5 & 13.1 & 26.0 & n/m & 2.4 & 6.0\\
PyTorch BF16 eager & 19.7 & 18.1 & 7.2 & 17.2 & 7.9 & 7.5 & 10.5\\
PyTorch BF16 compiled & 9.6 & 11.3 & 5.5 & 11.5 & 8.0 & 7.5 & 11.5\\
\grouprow{8}{OpenVLA-OFT}
AQ-engine F16 & 22.1 & 20.5 & 15.1 & 26.3 & 15.6 & 15.0 & 16.9\\
RTN Q4\_0 MMQ & n/m & 17.7 & 19.5 & 34.7 & n/m & 5.2 & 6.9\\
RTN Q4\_0 cuBLAS & 27.7 & 25.2 & 18.0 & 33.5 & 6.1 & 5.4 & 7.3\\
AWQ W4 MMQ & n/m & 18.7 & 20.3 & 36.7 & n/m & 5.6 & 7.4\\
AWQ W4 cuBLAS & 28.4 & 25.6 & 18.2 & 33.6 & 6.5 & 5.9 & 7.8\\
GPTQ W4 MMQ & n/m & 18.6 & 20.4 & 36.8 & n/m & 5.6 & 7.4\\
GPTQ W4 cuBLAS & 28.4 & 25.5 & 18.2 & 33.9 & 6.5 & 5.9 & 7.8\\
ActQuant 4\,bpw MMQ & 32.9 & 18.6 & 21.2 & 37.6 & 5.4 & 4.8 & 6.6\\
ActQuant 4\,bpw cuBLAS & n/m & 24.7 & 14.7 & 30.7 & n/m & 5.1 & 7.0\\
PyTorch BF16 & 18.1 & 15.7 & 10.4 & 19.7 & 15.9 & 15.4 & 17.9\\
\bottomrule
\end{tabular}
\end{table}

\begin{table}[!htbp]
\centering
\caption{GPU time per policy call by component for the official unquantized models (eager PyTorch, batch 1; ms, share of GPU time in parentheses), and end-to-end time.}
\label{tab:cross_e2}
\scriptsize
\begin{tabular}{@{}l l r r r@{}}
\toprule
Model & Component & H100 NVL & L4 & Jetson AGX Thor\\
\midrule
\pifive{} & SigLIP & 7.6 (13\%) & 18.4 (10\%) & 19.0 (11\%)\\
 & PaliGemma prefix & 13.2 (22\%) & 86.5 (48\%) & 71.4 (43\%)\\
 & action expert & 37.5 (63\%) & 75.1 (41\%) & 74.9 (45\%)\\
 & end to end (host overhead) & 130.5 (71.1) & 329.8 (148.8) & 232.2 (65.8)\\
OpenVLA-OFT & Llama-2 7B & 30.7 (76\%) & 172.0 (86\%) & 157.5 (86\%)\\
 & end to end & 59.3 & 217.4 & 181.7\\
GR00T N1.5 & DiT action head & 18.7 (58\%) & 55.4 (59\%) & 49.9 (54\%)\\
 & end to end & 114.1 & 127.3 & 109.5\\
UniVLA & Llama-2 7B & 43.8 (88\%) & 275.8 (95\%) & 265.1 (95\%)\\
 & end to end & 113.3 & 311.0 & 288.7\\
\bottomrule
\end{tabular}
\end{table}

\paragraph{Closed-loop energy on the Thor.} As an edge-device complement, every E3 configuration was also run alone on the Thor for the 10 LIBERO-Spatial episodes of trial 1, integrating board power over the whole run (server, policy and simulator). Per episode, \pifive{} F16 used 202\,J on the GPU rail (556\,J module), its W4 dequantize-path variants 220--246\,J and ActQuant 4\,bpw 346\,J; OpenVLA-OFT F16 249\,J and its W4 variants 276--318\,J; GR00T FP 130\,J and QuantVLA 182\,J. Quantization never lowered closed-loop energy on the edge device.
% =============================================================================

\end{document}